\documentclass[conference]{IEEEtran}
\IEEEoverridecommandlockouts
\usepackage{lcsys}
\usepackage{cite}
\usepackage{amsmath,amssymb,amsfonts}
\usepackage{algorithmic}
\usepackage{graphicx}
\usepackage{textcomp}

\usepackage{algorithmic}
\usepackage{graphicx}
\usepackage{textcomp}
\usepackage{xcolor}
\usepackage{caption, subcaption}
\usepackage{siunitx}
\usepackage{float}
\usepackage{gensymb}  % For degree symbol 
\usepackage{pstool,cleveref}
\usepackage{makecell}
\usepackage{yfonts}
\usepackage{multirow}
\usepackage{makecell}
\usepackage[inkscapelatex=false]{svg}
\usepackage{lipsum} % For dummy text
\usepackage{geometry}
\usepackage{bbm}
\usepackage{tabularx}
\usepackage{booktabs}
\usepackage{multicol}
\usepackage{cuted}

\def\BibTeX{{\rm B\kern-.05em{\sc i\kern-.025em b}\kern-.08em
    T\kern-.1667em\lower.7ex\hbox{E}\kern-.125emX}}
\begin{document}
\newgeometry{top=21.2mm, bottom=15.2mm, left=16.9mm, right=16.9mm}
\title{Quasi-Static Continuum Model of Octopus-Like Soft Robot Arm Under Water Actuated by Twisted and Coiled Artificial Muscles (TCAMs)}
\author{Amirreza Fahim Golestaneh$^{1}$, Venanzio Cichella$^{1}$ and Caterina Lamuta$^{1}$ 
%\thanks{This work was supported by NASA and ONR. }
\thanks{$^{1}$ The authors are with the Department of Mechanical Engineering, University of Iowa, Iowa City, IA 52240 {\tt\small \{amirreza-fahimgolestaneh, venanzio-cichella, caterina-lamuta\} @uiowa.edu} }
}
\maketitle
%
%%%%%%%%%%%% Abstract %%%%%%%%%%%%%%%%%%%%%%%%%%
\begin{abstract}
The current work is a qualitative study that aims to explore the implementation of Twisted and Coiled Artificial Muscles (TCAMs) for actuating and replicating the bending motion of an octopus-like soft robot arm underwater.
Additionally, it investigates the impact of hydrostatic and dynamic forces from steady-state fluid flow on the arm's motion.
The artificial muscles are lightweight and low-cost actuators that generate a high power-to-weight ratio, producing tensile force up to 12,600 times their own weight, which is close to the functionality of biological muscles.
The "extended" Cosserat theory of rods is employed to formulate a quasi-static continuum model of arm motion, where the arm's cross-section is not only capable of rigid rotation but also deforms within its plane.
This planar deformation of the arm cross-section aligns with the biological behavior of the octopus arm, where the stiffness of the hydrostat is directly induced by the incompressibility of the tissues.
%Here, the derivation of the model is discussed in detail, specifically the impacts of hydrostatic and dynamic forces of steady-state fluid flow on the motion.
%Unlike the standard Cosserat theory, a new cross-sectional normal strain is introduced to describe the deformation within the plane of the cross-section.
%The cross-section of the robot arm is not constrained to be locally normal to the direction of the rod at any point along its length.
%This is, in fact, of significant importance for modeling hollow rods and rods with large bending deformations, such as the octopus arm.
In line with the main goal, a constitutive model is derived for the material of the octopus arm to capture its characteristic behavior. 
\end{abstract} 
\begin{IEEEkeywords}
Extended Cosserat Theory of Rods, Continuum Soft Robots, Twisted and Coiled Artificial Muscles (TCAMs), Octopus Arm Modeling, Fluid Flow Forces, Constitutive modeling.
\end{IEEEkeywords}
%
%%%%%%%%%%%%%%% SECTION %%%%%%%%%%%%%%%%%%%%%%%%
\section{Introduction}
\label{Sec_Introduction}
Early work on robot manipulation focused on discrete rigid manipulators, crafted to replicate the motion and grasping of the human hand \cite{Gianpaolo2020_a}. 
These robots consist of rigid links connected by joints, culminating in an end-effector that interacts with the environment. 
Therefore, each degree of freedom (DOF) relies on an actuatable joint, which limits the system's agility. 
Consequently, enhancing maneuverability requires adding more joints and DOFs, increasing complexity in both design and control.
Besides \textit{discrete} robots, Robinson et al. \cite{Robinson1999_a} categorized them into \textit{serpentine} robots (also known as \textit{hyper-redundant} manipulators) and \textit{continuum} robots. 
Hyper-redundant manipulators typically feature a backbone structure with numerous actuatable joints, resulting in a DOF significantly greater than that of their operational workspace.
For more details on modeling and solving the kinematics of hyper-redundant manipulators see early works \cite{ChirikjianPhD1992,Chirikjian1994_a,Chirikjian1995_a}. 
They also employed the continuum formulation of kinematics to optimize the configuration of these manipulators \cite{Chirikjian1995_a}. 
%They formulated the weak forms of the governing differential equations and applied the calculus of variation. 
A continuum robot is a type of manipulator that can locally change its configuration at any point along its length \cite{Trivedi2008_a}. 
This allows it to navigate workspaces with obstacles and reach almost any point within them. 
Resembling the movements of snakes \cite{Qin2022a}, elephant trunks \cite{Hannan2003_a}, and octopus arms \cite{Laschi2009a}, continuum robots theoretically possess infinite DOFs. 
However, unlike other manipulators, not every DOF is actuated, resulting in a relatively small number of actuators generating the motion \cite{Hannan2003_a}. 
These manipulators are typically actuated by cables, hydraulic, or pneumatic pressures \cite{Renda2012_a,Marchese2015_a,Rus2015_a}. \par
The analysis of motion in continuum robots is categorized into piecewise constant-field techniques \cite{Hannan2003_a, Katzschmann2019_a, DellaSantina2018_a, Renda2018_a,Sayadi2024_a} and continuous-field techniques \cite{Renda2020_a, Boyer2021_a, Xun2024_a, Roshanfar2024_a, Doroudchi2021_a, Till2019_a}.
One commonly adopted model for multi-sectional continuum robots is the piecewise constant curvature model, where it is assumed that the curvature remains constant along each section of the robot arm \cite{Hannan2003_a, Katzschmann2019_a, DellaSantina2018_a,Sayadi2024_a}.
The constant curvature technique becomes inaccurate in many scenarios, such as when significant external loads, like gravity, are present, or when there is considerable variation locally in the curvature of the robot arm \cite{Camarillo2008_a,Tatlicioglu2007_a}.
For more details on piecewise constant curvature model see \cite{Webster2010_a}.
%Katzschmann et al. \cite{Katzschmann2019_a} used the piecewise constant curvature approach to dynamically analyze and control the motion of a multi-segment soft robot, as a continuum manipulator, in three-dimensional ($3$D) Euclidean space $\mathbb{R}^3$, that is the extension of their previous work \cite{DellaSantina2018_a} in $2$D space.
\par
The \textit{standard} Cosserat theory of rods \cite{Antman_Book}, widely adopted as a continuous-field technique, offers an accurate formulation for the motion of continuum robots \cite{Renda2020_a, Boyer2021_a, Xun2024_a, Roshanfar2024_a, Doroudchi2021_a, Till2019_a}. 
This approach utilizes strain fields corresponding to the rod's deformation to derive the dynamic motion of the robot arm.
Previous works \cite{Renda2012_c,Renda2018_a} explains how standard Cosserat theory is implemented to model a cable-driven soft robot arm, followed by the application of the piecewise constant-strain technique for discretization \cite{Renda2018_a}. 
This approach was further developed in \cite{Renda2020_a} to incorporate a variable-strain methodology.
Similarly, in \cite{Boyer2021_a}, a strain-based technique was employed to construct a nonlinear parametrization of the Cosserat model for the soft robot arm and to conduct dynamic motion analysis.
Unlike the standard Cosserat theory, which focuses solely on the deformation of the entire rod, the \textit{extended Cosserat} theory of rods also incorporates planar deformation within the rod's cross-section alongside the rod's overall deformation \cite{Kumar2010_a}. 
This implies that, the cross-section of the rod can undergo both rigid rotation and deformation within its plane.
Incorporating the deformation in rod cross section can inherently enhance the dynamic analysis of the robot arm motion particularly for arms that undergo substantial deformations.  
For more details on application of Cosserat rod theory in modeling of continuum robots, see reviews \cite{Qin2024a, Armanini2023a}. 
\par
The recently developed Twisted and Coiled Artificial Muscles (TCAMs) offer lightweight and cost-effective alternatives to traditional actuators, such as electromagnetic motors, and hydraulic and pneumatic actuators \cite{Leng2021_a, Yang2017_a, Tawfick2019_a}. 
These artificial muscles generate a high power-to-weight ratio, producing tensile forces up to 12,600 times their own weight, closely resembling the functionality of biological muscles \cite{Saharan2019_a, Kotak2022_a}.
%TCAMs can be fabricated from diverse components such as carbon nanotubes embedded in paraffin wax \cite{Lamuta2019a}, polymer fishing lines \cite{Lamuta2019a, Haines2014a}, or carbon fibers combined with silicone rubber \cite{Lamuta2018a}. 
%These artificial muscles are created by coating fibers with a thermally active polymer, twisting them into helical structures, and heat-treating them to set their shape \cite{Bell2022_a}. 
%This process enables them to contract when activated by stimuli like electric currents or temperature changes.
The integration of TCAM actuators in soft robotics is rapidly growing due to their high power-to-weight ratio and low manufacturing cost \cite{Saharan2019_a}. 
This application necessitates a thorough understanding and the development of a theoretical dynamic model of TCAMs to accurately describe their actuation behavior and facilitate the design of control algorithms. 
%Despite previous theoretical modeling efforts on TCAM dynamics and control \cite{Tsabedze2021a, Wu2020a}, the behavior of TCAMs remains insufficiently explored.
As discussed in \cite{Weerakkody2023a}, TCAM dynamics are categorized into fitting-based models, which derive constant parameters from experimental data, and physics-based models, utilizing constitutive relations. 
Weerakkody et al. \cite{Weerakkody2023a} developed a physics-based model akin to Giovinco et al. \cite{Giovinco2020a} for TCAM actuation, introducing an adaptive control algorithm. 
This model extends their earlier static work in \cite{Lamuta2018a}, comprising a thermal model linking input voltage or heat to fiber radius expansion in TCAMs, and a mechanical model describing how this radius expansion relates to generated tensile force in TCAMs.
For more details on physics-based modeling of artificial muscles, refer to \cite{Giovinco2020a, Lamuta2018a, Hammond2022a, Sun2022a}. \par
The primary goal of this work is to study the impact of hydrostatic and dynamic forces from steady-state fluid flow on the motion of a robot arm driven by TCAMs. 
Throughout this paper, the extended Cosserat theory of rods models an octopus-like soft robot arm as a continuum robot, allowing for both rigid rotation and planar deformation of the rod's cross-section.
This implies that the cross-section is not locally constrained to be perpendicular to the rod's direction at any point along its length, which is crucial for modeling hollow rods and those with significant bending deformations, such as octopus arms.
The planar deformation of the cross-section is aligned with the biological behavior of octopus arms, where the stiffness of the hydrostat is induced by tissue incompressibility.
The study explores the application of recently developed electro-thermo TCAMs, developed by some of the authors, to actuate and replicate the motion of octopus arms. 
A physics-based dynamic model of TCAMs is discussed to characterize their time-varying tensile actuation. 
Furthermore, a constitutive model is derived for the material of the octopus arm to capture its characteristic behavior. %, aligning with the primary objective of the study.
%
%For the rest of this paper, the standard and extended Cosserat theory of rods are discussed in Section~\ref{}. 
%The thermal and mechanical actuation models of the TCAMs are explained in Section~\ref{}. 
%Section~\ref{} formulates the cable-driven forces and hydrostatic and dynamic forces of the fluid flow exerted on the robot arm. 
%The constitutive model for the robot arm material is derived in Section~\ref{}. 
%Section~\ref{} presents the governing equations of equilibrium that is followed by the local volume preservation constraint. 
%The results are discussed in Section~\ref{} and concluded in Section~\ref{}.
%
%%%%%%%%%%%% Standard Cosserat Theory of Rods %%%%%%%%%%%%%
\section{Extended Cosserat Theory of Rods} 
\label{Sec_StandardCosseratTheoryOfRods}
\subsection{Configuration of Rod}
\label{Sec_StandardConfigurationOfRod}
\begin{figure}
% Include 1st image.
\begin{subfigure} {0.5 \textwidth}
\centering
\centerline{\includegraphics[height= 1.2 in]{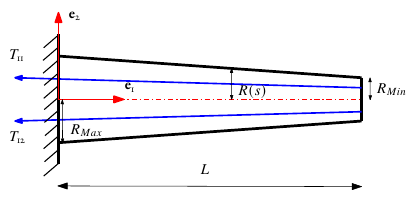} }
\caption{Reference (undeformed) straight arm.} 
\label{Fig_UnDefArm}
\end{subfigure} \vspace{0.1 in}
%
%\newline
% Include 2nd image
\begin{subfigure}{0.5\textwidth}
\centering
\centerline{\includegraphics[height= 2.2 in]{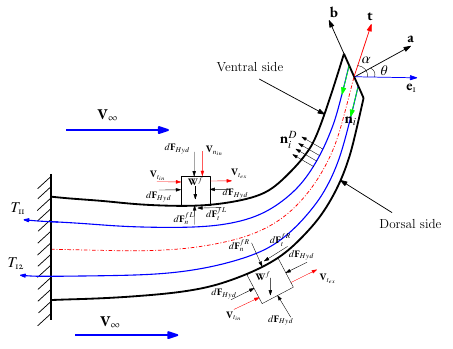}}
\caption{Deformed arm.} 
\label{Fig_DefArm}
\end{subfigure}  \vspace{0.1 in}
\caption{Octopus arm actuated by two TCAMs.}
\label{Fig_UndefDefArm}
\end{figure}

Fig.~\ref{Fig_UndefDefArm} shows both the reference (undeformed) straight configuration and the deformed configuration of the octopus-like arm, actuated by two longitudinal TCAMs.   
%Two infinitesimal control volume elements are considered on both sides of the deformed configuration in Figure~\ref{Fig_DefArm} to study the impacts of the steady state flow of fluid over the arm.
The \textit{extended} Cosserat theory of rods \cite{Kumar2010_a} is used to model the configuration of the robot arm. 
This theory retains the standard Cosserat framework, including all strains and governing equations that describe the bulk rod deformation, while adding new strains with corresponding equations to account for cross-sectional deformation. 
For a two-dimensional (2D) rod, this involves introducing a new normal strain $\beta \in \mathbb{R}$ to represent the planar deformation of the cross section, followed by the rigid rotation $\mathbf{R} \in \mathbb{SO}(2)$ to determine the orientation of the cross section. 
The configuration of the robot arm, $\mathbf{x}: [0, L] \longmapsto \mathbb{R}^2$, is parametrized by the curve length $s \in [0, L]$ of the centerline of the undeformed straight rod: 
\begin{align}
%& \mathbf{x}: [0, L]  \longmapsto \mathbb{R}^2 \nonumber \\[5pt]
& \mathbf{x}(s)= \mathbf{r}(s) + x_2(X_2(s), s) \mathbf{b}(s), 
\label{Eq_DeformedConfig}
\end{align}  
where $\mathbf{r}(s)$ represents the centerline of the rod,  $x_2(X_2(s), s) \doteq \beta(s) X_2(s)$ such that $\left \vert X_2(s) \right \vert$ is the radius of undeformed rod and $\beta(s)= \frac{\partial x_2(s)}{\partial X_2(s)}$ accounts for the planar deformation of the cross section along $\mathbf{b}$. 
The standard orthonormal \textit{directors} $\mathbf{a}(s)= \cos(\theta) \mathbf{e}_1 + \sin(\theta) \mathbf{e}_2$ and $\mathbf{b}(s)= -\sin(\theta) \mathbf{e}_1 + \cos(\theta) \mathbf{e}_2$ determine the rotation matrix $\mathbf{R}= \begin{bmatrix} \mathbf{a} & \mathbf{b} \end{bmatrix}$, where $\{ \mathbf{e}_1, \mathbf{e}_2, \mathbf{k} \}$ form the global fixed Cartesian frame located on the centerline at $s= 0$ and angle  $\theta(s): [ 0,L ]  \longmapsto \mathbb{R}$ defines the orientation of $\mathbf{a}$ (see Fig.~\ref{Fig_DefArm}).
%   
%%%%%%%%%%%%%%%%%%%%%%%%%%%%%%%%%%%%%%
%
%%%%%%%%%%%%%%%%%%%%%%%%%%%%%%%%%%%%%%
\subsection{Bulk Strains}
\label{Sec_Bulk Strains}
In standard Cosserat theory of rods, the bulk translational strains, which correspond to the deformation of the entire rod, are defined as the components of $\mathbf{r}_s \in \mathbb{R}^2$ represented in the standard body frame $ \{ \mathbf{a}, \mathbf{b} \}$ 
\begin{align}
%& \mathbf{t}: [0, L]  \longmapsto \mathbb{R}^2 \nonumber \\[5pt]
& \mathbf{t}(s) \doteq \mathbf{r}_s(s)= \nu(s) \mathbf{a}(s) + \eta(s) \mathbf{b}(s),  
\label{Eq_SpatialDerivative_rVector}
\end{align}  
where $\nu \in \mathbb{R}$ and $\eta \in \mathbb{R}$ respectively measure the normal and shear translational bulk strains and subscript $s$ signifies spatial derivative $( )_s \doteq \partial_s ( )$. 
Similarly, the bulk bending strains are defined as the components of $\mathbf{a}_s$ and $\mathbf{b}_s$, expressed in the standard body frame:
\begin{align}
&\mathbf{a}_s(s)= R_s(s) \mathbf{e}_1= \mu \mathbf{b}, \nonumber \\[5pt]
&\mathbf{b}_s(s)= R_s(s) \mathbf{e}_2= - \mu \mathbf{a}, 
\label{Eq_SpatialDerivative_ABDirectors}
\end{align}  
where the bending strain $\mu(s)= \partial_s \theta(s) \in \mathbb{R}$ locally measures the curvature of the centerline. 
The skew-symmetric matrix $\xi(s) \doteq  \mathbf{g}^{-1} \mathbf{g}_s= \begin{bmatrix} \mathbf{R}^T \mathbf{R}_s & \mathbf{R}^T \mathbf{r}_s \\ 0 & 0 \end{bmatrix}  \in \mathbbm{se}(2)$ captures the bulk strains, where $\mathbf{g}(s) \doteq \begin{bmatrix} \mathbf{R} & \mathbf{r} \\ 0 & 1 \end{bmatrix} \in \mathbb{SE}(2)$ is the rigid body transformation. 
The centerline $\mathbf{r}$ is calculated from strains by:
\begin{align}
 %r^a_s, r^b_s &: [0, L] \longmapsto \mathbb{R} \nonumber \\[5pt]
& r^a_s(s)= \nu + \mu r^b,   \nonumber \\[5pt]
& r^b_s(s)= \eta - \mu r^a, 
\label{Eq_rarbSpaceCenterlineComponents}
\end{align}
where $r^a$ and $r^b $ are its coordinates in standard body frame. 
%%%%%%%%%%%% TCAMs %%%%%%%%%%%%%%%%%%%%%%%%%%
\section{TCAMs Actuation Model}
\label{Sec_TCAMs Actuation Model}
This section reviews the theory behind the recently developed electro-thermo TCAMs, by some of the authors \cite{Weerakkody2023a}. %for actuating and replicating the bending motion of an octopus-like soft robot arm underwater.
The theory consists of a thermal model and a mechanical model, which together describe the time-varying tensile actuation of TCAMs. 
The thermal model details the relationship between the input voltage or heat and the increase in the fiber radius in electro-thermally actuated TCAMs. 
The mechanical model relates the increase in fiber radius to the generated tensile force in the TCAMs. 
%Notably, this is a physics-based model where the dynamic behavior of TCAM actuation is mathematically defined by mechanical, electro-thermal, and material characteristics through constitutive relations. 
%This implies that the model can be generally applied to TCAMs made from various fiber materials and matrices, provided the appropriate constitutive constants are used.
%
%%%%%%%%%%%% Thermal Model %%%%%%%%%%%%%%%%%%%%%%%%%%
\subsection{Thermal Model For TCAMs}
\label{Sec_ThermalModelForTCAMs}
TCAMs are manufactured by coating fibers with a thermally active polymer, then twisting them until they spontaneously coil. 
They undergo precise heat treatment to enable contraction when activated by electro-thermal stimuli. 
This contraction is caused by anisotropic volume expansion of the TCAM fibers after stimulation. 
The fibers' anisotropic thermal properties cause their radius to increase upon exposure to heat, Joule heating, or absorption of a chemical solvent, while maintaining a constant length. 
For some materials like nylons, the fiber length shortens while the radius increases after stimulation \cite{Haines2014a}. 
The increase in fiber radius $r_{tm} \in \mathbb{R}^+$ can be linearly correlated with the rise in temperature: 
\begin{align}
 \frac{r_{tm}(\bar{T}_{\Delta}) - r^o_{tm}}{r^o_{tm}}= \alpha \, \bar{T}_{\Delta}(t), 
\label{Eq_TCAMsRadiusTemperature}
\end{align}
where $\alpha \in \mathbb{R}^+$ is the linear thermal expansion coefficient in the radial direction, $r^o_{tm} \in \mathbb{R}^+$ denotes the initial fiber radius and $\bar{T}_{\Delta}= \bar{T}(t) - \bar{T}_{amb} \in \mathbb{R}$ is the change in temperature such that $\bar{T}_{amb} \in \mathbb{R}^+$ represents the ambient temperature. 
When the TCAMs are electro-thermally stimulated, the relationship between the input voltage $v_{vol}(t) \in \mathbb{R}$ and the change in temperature $\bar{T}_{\Delta} \in \mathbb{R}$ can be defined by \cite{Yip2015a}:
\begin{align}
%\bar{T}_{\Delta} &: \mathbb{R}^{\geq 0} \longmapsto \mathbb{R} \nonumber \\[5pt]
 \dot{\bar{T}}_{\Delta}(t) &= \frac{v_{vol}^2(t)}{C_t R_{vol}} - \frac{\lambda(\bar{T}_{\Delta})}{C_t} \bar{T}_{\Delta}, 
\label{Eq_TCAMsVoltageTemperature}
\end{align}
where $\dot{( \, )}$ represents time derivative and $C_t= m c_p \in \mathbb{R}^+$, $R_{vol} \in \mathbb{R}^+$ and $\gamma= h A_s \in \mathbb{R}^+$ are respectively the thermal mass, electrical resistance, and the absolute thermal conductivity of the TCAM. 
Here, $m \in \mathbb{R}^+$ is the mass of the TCAM, $A_s \in \mathbb{R}^+$ is its cross section, $h \in \mathbb{R}^+$ is the convection coefficient between the actuator and the environment, and $c_p \in \mathbb{R}^+$ is the TCAM specific heat. \par
%
%%%%%%%%%%%% Mechanical Model %%%%%%%%%%%%%%%%%%%%%%%%%%
\subsection{Mechanical Model For TCAMs}
\label{Sec_MechanicalModelForTCAMs}
The increase in the radius of TCAM fibers and the consequent anisotropic volume expansion of the fibers stiffen the bending and torsional deformations of the TCAMs, such that:
\begin{align}
%B, C &: \mathbb{R} \longmapsto \mathbb{R}^+ \nonumber \\[5pt]
B(\bar{T}_{\Delta}) &= E_x \frac{\pi}{4} r^4_{tm}(\bar{T}_{\Delta}), \nonumber \\[5pt]
C(\bar{T}_{\Delta}) &= G_{yz} \frac{\pi}{2} r^4_{tm}(\bar{T}_{\Delta}),
\label{Eq_TCAMsBendingTorsionStiffness}
\end{align}
where $E_x, G_{yz} \in \mathbb{R}^+$ are respectively normal Young's modulus and shear modulus of TCAMs. 
The variation in bending and torsion stiffness of the TCAMs causes the change in the TCAMs geometry of coil angle $\beta(\bar{T}_{\Delta}) \in \mathbb{R}$ 
 \begin{align}
 %\cos( \beta(\bar{T}_{\Delta})) &: \mathbb{R} \longmapsto [-1,1] \nonumber \\[5pt]
 \cos( \beta(\bar{T}_{\Delta})) = 1- \frac{ \left(  \frac{q B(\bar{T}_{\Delta}) C(\bar{T}_{\Delta})}{\sqrt{B(\bar{T}_{\Delta})} \bar{m} g^r} - 2 C(\bar{T}_{\Delta} ) \right)}{B(\bar{T}_{\Delta}) - C(\bar{T}_{\Delta})}, 
\label{Eq_TCAMsCoilAngleBeta}
\end{align}
where $q= \frac{2 \pi n}{\bar{S}} \in \mathbb{R}^+$ denotes the total end rotation of the fiber, $n \in \mathbb{R}^+$ is the number of turns and $\bar{m} \in \mathbb{R}^+$ denotes the external mass applied to the end of the TCAMs during the coiling process (the TCAM mass is neglected compared to mass $\bar{m}$).
Here, $g^r \in \mathbb{R}^+$ is the magnitude of gravity acceleration.
The length of the TCAM after contraction is then calculated as:
\begin{align}
%\mathcal{L} &: \mathbb{R}^{\geq 0} \longmapsto \mathbb{R}^+ \nonumber \\[5pt]
\mathcal{L}(t) &= \bar{S} \cos( \beta(\bar{T}_{\Delta})).
\label{Eq_TCAMsLength}
\end{align}
The dynamics of TCAMs actuation is modeled as a second-order mass-spring-damper system, governed by: 
\begin{align}
m \mathcal{\ddot{L}}(t) + b(\bar{T}_{\Delta}(t)) \mathcal{\dot{L}}(t) + k(\bar{T}_{\Delta}(t)) \mathcal{L}(t) = F_e(t), 
\label{Eq_TCAMsDynamicEquationOfMotion}  
\end{align}
where $b(\bar{T}_{\Delta}(t)) \in \mathbb{R}^+$ and $k(\bar{T}_{\Delta}(t)) \in \mathbb{R}^+$ and $F_e(t) \in \mathbb{R}$ are respectively the time-dependent damping coefficient, the spring coefficient of the TCAMs and the external force applied to the system. 
Considering an equilibrium configuration such that $F_e(t)= \bar{m} g$, where $\mathcal{\ddot{L}}(t) = \mathcal{\dot{L}}(t)= 0$, then the time-dependent spring coefficient of the TCAMs is obtained: 
\begin{align}
%k &: \mathbb{R} \longmapsto \mathbb{R}^+ \nonumber \\[5pt]
k(\bar{T}_{\Delta}) &= \frac{\bar{m} g}{\bar{S} \cos( \beta(\bar{T}_{\Delta}))}.  
\label{Eq_TCAMsSpringStiffnessK}
\end{align}
The time-dependent damping coefficient $b(\bar{T}_{\Delta}) \in \mathbb{R}^+$ of the TCAMs is calculated in correspondence to the viscosity coefficient $\eta(\bar{T}_{\Delta}) \in \mathbb{R}^+$ of the TCAMs, where the Kelvin-Voigt viscoelastic theory is used to describe the constitutive behavior of the TCAMs as a viscoelastic material \cite{Kato2018a}
\begin{align}
%b &: \mathbb{R} \longmapsto \mathbb{R}^+ \nonumber \\[5pt]
b(\bar{T}_{\Delta}) &= \frac{\eta(\bar{T}_{\Delta}) A_s}{\mathcal{\bar{L}}}. 
\label{Eq_TCAMsViscosityCoefficient}  
\end{align}
Now let define the TCAMs contraction as $d(t)= \mathcal{L}_o - \mathcal{L}(t)$, where the initial length of the TCAMs $\mathcal{L}_o > 0$ at time $t= 0$ for ambient temperature of TCAMs $\bar{T}(t=0)= \bar{T}_{amb} $ is calculated as $\mathcal{L}_o= \frac{\bar{m}g}{k_o}$, where $k_o \doteq k(\bar{T}_{\Delta}(t=0)=0)$.
The dynamic equation of motion of TCAMs \eqref{Eq_TCAMsDynamicEquationOfMotion} is reformulated: 
\begin{align}
m \ddot{d}(t) + b(\bar{T}_{\Delta}(t)) \dot{d}(t) &+ k_o d(t)= \nonumber \\[5pt] 
& k(\bar{T}_{\Delta}(t)) \mathcal{L}_o - F_e - k_{\Delta}(t) d(t), 
\label{Eq_TCAMsDynamicEquationOfMotionContraction}  
\end{align}
where $k_{\Delta}(t) \doteq k(\bar{T}_{\Delta}(t)) - k_o$. 
Linearize \eqref{Eq_TCAMsDynamicEquationOfMotionContraction} by considering $C \bar{T}_\Delta(t) \approx k(\bar{T}_{\Delta}(t)) \mathcal{L}_o - F_e - k_{\Delta}(t) d(t)$, where the coefficient $C$ can be experimentally approximated by some arbitrary constant temperature $\bar{T}^\ast_\Delta $. 
Consider the equilibrium condition from \eqref{Eq_TCAMsDynamicEquationOfMotionContraction} at an arbitrary time $t^\ast > 0$ with arbitrary temperature $\bar{T}^\ast $ and $\bar{T}^\ast_\Delta(t^\ast)= \bar{T}^\ast(t^\ast) - \bar{T}_{amb}$, where $\mathcal{\dot{L}}(t^\ast)= \mathcal{\ddot{L}}(t^\ast)= 0$: 
\begin{align}
    k_o d^\ast(t^\ast) \approx C T^\ast_\Delta(t^\ast), 
    \label{Eq_EquilibriumEquationTCAMs}
\end{align}
where $d^\ast(t^\ast)= \mathcal{L}_o - \frac{mg}{k(\bar{T}^\ast_\Delta(t^\ast))}$ such that $k(\bar{T}^\ast_\Delta(t^\ast))$ is calculated from \eqref{Eq_TCAMsSpringStiffnessK}. 
Therefore, the coefficient $C$ is:
\begin{align}
    C= \frac{k_o d^\ast(t^\ast)}{\bar{T}^\ast_\Delta(t^\ast)}
    \label{Eq_CoefficentC_TCAMs}
\end{align}
and the dynamic equation of motion of the TCAMs is approximated as:
\begin{align}
    m \ddot{d}(t) + b(\bar{T}_{\Delta}(t)) \dot{d}(t) + k_o d(t) \approx \frac{k_o d^\ast(t^\ast)}{\bar{T}^\ast_\Delta(t^\ast)} \bar{T}_\Delta(t),
    \label{Eq_TCAMsDynamicEquationOfMotionLinearized}
\end{align}
where $\bar{T}_\Delta(t)$ is associated to the applied input voltage $V(t) \in \mathbb{R}^+ $ through the ODE in \eqref{Eq_TCAMsVoltageTemperature}. \\
%
%%%%%%%%%%%%% External Forces and Moments %%%%%%%%%%%%%%
\section{External and Internal Loads}
\label{Sec_ExternalForcesAndMoments}
\subsection{Concentrated Force of TCAMs}
\label{Sec_ConcentratedForceOfTCAMs}
The concentrated force $\mathbf{f}^{cc}_{i j}(L_i) \in \mathbb{R}^2$ of the TCAMs $j \in \{ 1, 2\}$ of the segment $i \in \mathbb{Z}^+$ is modeled as \cite{Renda2014_a}:
\begin{align}
& \mathbf{f}_{ij}^{cc}(L_i)= - T_{ij} \mathbf{t}_{ij}^c (L_i), 
\label{Eq_ConcentratedTCAMsForce}
\end{align} 
where $ L_i  \in (0, L] $ is the length of undeformed centerline of the segment $i \in \mathbb{Z}^+$ and $T_{i j} \in \mathbb{R}^+ $ denotes the tension of the corresponding TCAMs and segment
\begin{align}
    T_{i j}(\bar{t})= \frac{k_o d^\ast(t^\ast)}{\bar{T}^\ast_\Delta(t^\ast)} \bar{T}^{ij}_\Delta(\bar{t}).
    \label{Eq_CableTensionTCAMs}
\end{align}
In this context, the segment refers to an individual, flexible section of the arm that is actuated independently, allowing fine control over complex shapes. 
%\textcolor{red}{In this context, the segment refers to an individual, flexible section of the arm that is actuated. 
%Each segment can deform independently, allowing fine control over complex shapes. }
The temperature $\bar{T}^{ij}_\Delta(\bar{t})$ at any fixed time $\bar{t} \in \mathbb{R}^+$ is associated to the applied input voltage $V^{ij}(\bar{t}) \in \mathbb{R}^+ $ through the ODE in \eqref{Eq_TCAMsVoltageTemperature}.
The $ \mathbf{t}_{ij}^c (s)= \frac{\partial \mathbf{r}^c_{ij} (s)}{\partial s}$ for $ s \in [0,  L_i] $ is the tangent vector to the TCAMs, such that the position vector $\mathbf{r}_{ij}^{c}(s) \in \mathbb{R}^2$ of the TCAMs $j \in \{ 1, 2\}$ of the segment $i \in \mathbb{Z}^+$ is determined by $\mathbf{r}_{ij}^{c}(s)= \mathbf{r}(s) + y_{ij}^c (s) \beta(s) \mathbf{b}(s)$, where $y_{ij}^{c}(s) \in \mathbb{R}$ denotes the distance of that TCAMs from the undeformed centerline. 
%
%%%%%%%%%%%%% Distributed Force of TCAMs %%%%%%%%%%%%%%
\subsection{Distributed Force of TCAMs}
\label{Sec_DistributedForceOfTCAMs}
The distributed force per unit length $\mathbf{f}_{ij}^{cd}(s) \in \mathbb{R}^2$ of the TCAMs $j \in \{1, 2\}$ along the undeformed centerline $s \in [0, L_i]$ of segment $i \in \mathbb{Z}^+$ is modeled as \cite{Renda2014_a}:
\begin{align}
%& \mathbf{f}_{ij}^{cd}:  [0,  L_i]   \longmapsto \mathbb{R}^2 \nonumber \\[5pt]
& \mathbf{f}_{ij}^{cd}(s)= T_{ij} \frac{ \partial \mathbf{t}_{ij}^c (s)}{\partial s}. 
\label{Eq_DistributedTCAMsForcePerLength}
\end{align} 
Therefore, the moment $\mathbf{m}_{ij}^{cd}(s) \in \mathbb{R}^2$ about the origin of the fixed global Cartesian frame is:
\begin{align}
 %\mathbf{m}_{ij}^{cd}: &  [0,  L_i]   \longmapsto \mathbb{R}^2 \nonumber \\[5pt]
 \mathbf{m}_{ij}^{cd} &(s)=  \int_s^{L_i} \mathbf{r}_{ij}^c(\delta) \times  T_{ij} \frac{ \partial \mathbf{t}_{ij}^c (\delta)}{\partial \delta} \text{d} \delta.
\label{Eq_DistributedTCAMsMomentIntegral}
\end{align} 
%
%%%%%%%%%%%%%%%% Fluid Forces %%%%%%%%%%%%%%%%%%%
\subsection{ Fluid Forces}
\label{Sec_FluidForces}
The primary objective is to investigate how hydrostatic and dynamic forces from a steady-state fluid flow affect the bending motion of an octopus-like robot arm propelled by TCAMs. 
The study focuses on analyzing the static deformation of the arm as the tension in the TCAMs incrementally increases.
To achieve this goal, a viscous fluid flow with a quadratic velocity profile $\mathbf{v}_t(s,y)$ is introduced over the arm.
This velocity profile varies with the normal distance $y \in [0,h_\infty]$ from the arm's surface, satisfying the no-slip condition on the arm's surface. 
The velocity monotonically increases to its maximum $v_{t_\infty}$ at a distance $h_\infty$ from the arm's surface
\begin{equation}
\mathbf{v}_t(s,y)= v_{t \infty}(s) \left[  \frac{2 y}{h_\infty} - \left( \frac{y}{h_\infty} \right)^2  \right] \mathbf{u}_t (s).
\label{Eq_VelocityProfile}
\end{equation}  
The subscript $t$ signifies the tangential direction to the surface of the arm such that $\mathbf{u}_t (s)=  \frac{ \partial_s \mathbf{x}(s) }{ \Vert \partial_s \mathbf{x}(s) \Vert}$ is the tangent unit vector to the arm surface and $v_{t \infty}(s)= \mathbf{v}_{\infty} \cdot \mathbf{u}_t (s)$ is the tangential component of the free-stream velocity $\mathbf{v}_{\infty}= v_{\infty} \mathbf{e}_1$ of the fluid flow.  
The quadratic velocity profile $\mathbf{v}_t$ in \eqref{Eq_VelocityProfile} is used to calculate the viscus frictional force $\text{d}\mathbf{F}_{vis}(s) \in \mathbb{R}^2$ applied to the infinitesimal surface of the arm opposite to the direction of the flow:
\begin{align}
%& \mathbf{F}_{vis}:  [0,  L]   \longmapsto \mathbb{R}^2 \nonumber \\[5pt]
& \text{d} \mathbf{F}_{vis}(s)= \frac{- 2 \pi X_2(s) \beta \mu_v v_{t \infty}(s) }{ h_{\infty} } \mathbf{u}_t(s) \text{d}s,
\label{Eq_FrictionForce_dFvis}
\end{align}
where $\mu_v$ is the dynamic viscosity of the fluid.
The continuity equation and the principle of linear momentum for a steady state flow is applied separately over two infinitesimal control volumes of fluid on ventral and dorsal surfaces of the arm to compute the tangential $\text{d} \mathbf{F}_t^{V}(s)$ and normal $\text{d} \mathbf{F}_n^{V}(s), \, \text{d} \mathbf{F}_n^{D}(s) $ hydrostatic and dynamic contact forces applied by the fluid to the ventral and dorsal infinitesimal surfaces of the arm:
\begin{align}
\text{d} \mathbf{F}_t^{V}(s) =  
 \left( v^2_{n_{in}} (s)  +  2  v_{t_{in}}(s) v_{n_{in}}(s)  \right) \pi \rho_w  \beta X_2  \text{d} s \, \mathbf{u}_t(s),
\label{Eq_ForceLeftTangentRod}
\end{align}
\begin{align}
\text{d} \mathbf{F}_n^{V}(s) =
 \left[ \text{d} \mathbf{F}_{Hyd}(s) +  \pi \rho_w  \beta X_2 v^2_{n_{in}} (s)  \text{d} s  \right] \mathbf{u}_n(s),
\label{Eq_ForceLeftNormalRod}
\end{align}
\begin{align}
\text{d} \mathbf{F}_n^{D}(s) =
  \text{d} \mathbf{F}_{Hyd}(s)  \mathbf{u}_n(s). \hspace{1.3 in}
\label{Eq_ForceRightNormalRod}
 \end{align}   
$\mathbf{F}_{Hyd}(s)$ denotes the hydrostatic force of fluid, $\rho_w \in \mathbb{R}^+$ is the density of fluid, the superscripts $V$ and $D$ respectively signify the ventral and dorsal surfaces of the arm, subscript $n$ signifies the normal direction to the surface of the arm and subscript $in$ signifies the inlet open surface of the control volume and its corresponding quantities.
The weight of the infinitesimal control volume of the fluid is negligible compared to the other existing forces and therefore it is ignored in these computations. 
The normal component $v_n(s)$ of the free-stream velocity $\mathbf{v}_{\infty}= v_{\infty} \mathbf{e}_1$ is  $v_n(s)= \mathbf{v}_{\infty} \cdot \mathbf{u}_n (s)$, where $\mathbf{u}_n(s)= \mathbf{k} \times \mathbf{u}_t(s)$
such that $\Vert \mathbf{k} \times \mathbf{u}_t(s) \Vert= 1$ and $\{\mathbf{u}_t, \mathbf{u}_n, \mathbf{k} \}$ forms an orthonormal local body frame of reference. \par
The inlets and outlets of the control volume are determined in accordance with the direction of the velocity of the flow at each open surface of the control volume. 
The left and top surfaces of the control volume on the ventral surface of the arm serve as inlets, while the right surface is an outlet. 
Similarly, the left and right surfaces of the control volume on the dorsal surface of the arm are, respectively, an inlet and an outlet, with the flow through the top surface assumed to be negligible (see Figure~\ref{Fig_DefArm}).   \par
It is notable that the resultant normal force $\text{d} \mathbf{F}_n^{V}$ in \eqref{Eq_ForceLeftNormalRod}, applied to the infinitesimal ventral surface of the arm, comprises both the hydrostatic force of the fluid and the dynamic force generated by the change in velocity (linear momentum) of the fluid flow between the inlet and outlet of the control volume from $\mathbf{v}_{n_{in}}$ to $0$. 
The resultant normal force $\text{d} \mathbf{F}_n^{D}$ in \eqref{Eq_ForceRightNormalRod}, applied to the infinitesimal dorsal surface of the arm, comprises only the hydrostatic force of the fluid, as the fluid flow through the top surface of the control volume is assumed negligible.
The resultant tangential force $\text{d} \mathbf{F}_t^{V}$ in \eqref{Eq_ForceLeftTangentRod}, applied to the infinitesimal ventral surface of the arm, consists of the force generated by the change in velocity (linear momentum) of the fluid flow between the inlet and outlet of the control volume from $\mathbf{v}_{t_{in}}$ to $\mathbf{v}_{t_{in}} + \mathbf{v}_{n_{in}}$.
This indicates that the fluid flow exerts both hydrostatic and dynamic resultant forces on the surface of the arm, influencing the configuration of the deformed arm underwater.
%
%%%%%%%%%%%%%%%% Weight of Arm %%%%%%%%%%%%%%%%%%%
\subsection{ Weight of Segment of Arm }
\label{Sec_WeightOfSegmentOfArm}
In the deformed configuration of the arm, the cross section is oriented along $\mathbf{b}$, and $\mathbf{a}$ is not necessarily tangent to the centerline. 
Therefore, the weight of a segment of the deformed arm between $\delta \in [s, L]$ is calculated as:
\begin{align}
%& \mathbf{W}: [0,  L]   \longmapsto \mathbb{R}^2 \nonumber \\[5pt]
&\mathbf{W}(s)= - \pi \rho g^r \int_{\delta= s}^{L} \nu(\delta) \left\{ \beta(\delta) X_2(\delta) \right\}^2 \text{d} \delta \mathbf{e}_2, 
\label{Eq_WeightSegmentRod}
\end{align}
where $\rho \in \mathbb{R}^+$ denotes the density of the arm material.  
The normal strain $\nu$ appears in \eqref{Eq_WeightSegmentRod} to align the rod's cross section with the direction of its centerline. 
%
%%%%%%%%%%%%%%%% Contact Forces & Moments %%%%%%%%%%%%%%%%%%%
\subsection{Resultant Contact Force and Moment}
\label{Sec_ResultantContactForceMoment}
The constitutive model of the soft robot material connects the strains to the internal contact forces and moments.
Since the contact force is not always aligned with the normal unit director $\mathbf{a}$ of the deformed cross section, it is expressed in the local standard body frame:
\begin{align}
%& \mathbf{f}^c:  [0,  L]  \longmapsto \mathbb{R}^2 \nonumber \\[5pt]
&\mathbf{f}^c(s) \doteq \left\{ N(s) \mathbf{a}(s) + H(s) \mathbf{b}(s)  \right\} A(s),  
\label{Eq_BulkContactForceFcConstitutive}
\end{align} 
where $A(s) = \pi (\beta(s) X_2(s))^2$, and $N$ and $H$ represent the normal and shear forces per unit area (tractions) of the cross section, which correspond to the normal and shear strains $\nu$ and $\eta$, respectively.  
The constitutive model of the arm material defines the resultant contact moment in the local standard body frame as:
\begin{align}
%& \mathbf{m}^c:  [0,  L]  \longmapsto \mathbb{R}^2 \nonumber \\[5pt]
&\mathbf{m}^c(s) \doteq M(s) J_k(s) \mathbf{k},  
\label{Eq_BulkContactMomentMcConstitutive}
\end{align} 
where $M$ is proportional to the magnitude of the bending moment and corresponds to the bending strain (local curvature) $\mu$. 
$ J_k(s)= \frac{\pi ( \beta X_2)^4}{4} \in \mathbb{R}$ is the second moment of inertia of the circular cross section of the rod about $\mathbf{k}$ axis.  
%
%%%%%%%%%%%%%%%% Constitutive Model %%%%%%%%%%%%%%%%%%%
\section{Constitutive Model}
\label{Sec_Constitutive_Model}
The behavior of the soft material of the robot arm is described by a hyperelastic model.  
In a hyperelastic material, the stress field is a conservative function derived from a scalar-valued function called the strain energy function, by taking its derivative with respect to the strain field. 
Let's assume the strain energy function $\psi$ for the robot arm is a quadratic polynomial of bulk strains:
\begin{align}
 \psi (\nu, \eta, \mu, s)= &C_1(s) (\nu - \nu_o)^2 + \nonumber \\[5pt]
&C_2(s) (\eta - \eta_o)^2+ C_3(s) (\mu - \mu_o)^2,
\label{Eq_StrainEnergy}
\end{align} 
where $C_1(s)$, $C_2(s)$ and $C_3(s)$ respectively denote the normal, shear and bending stiffness coefficients of the arm material.  
%The strain energy function $\psi \in \mathbb{R}$ may explicitly depend on the spacial variable $s \in [0, L]$ to allow for variation of constitutive behavior of the elastic body within its spacial domain of the material, such that this dependency on the spacial variable $s \in [0, L]$ is absent in all the constitutive fields of the \textit{homogeneous } materials. 
%As mentioned above, for a hyperelastic material the normal, shear tractions and bending couple are respectively calculated from the derivatives of the strain energy function $\psi$ in \eqref{Eq_StrainEnergy} with respect to the normal $\nu(s)$, shear $\eta(s)$ and bending $\mu(s)$ bulk strain fields in \eqref{Eq_BulkBothStrainVectors}.
The constitutive equations for a continuum material must meet specific criteria to accurately model its behavior. 
The strain energy function needs to be sufficiently smooth as the quadratic polynomial in \eqref{Eq_StrainEnergy} is infinitely continuously differentiable $\psi (\nu, \eta, \mu, s) \in C^\infty( \mathbb{R}^3)$, assuming $\nu, \eta, \mu, C_i(s) \in C^\infty ( \mathbb{R})$ for $i \in { 1,2,3 }$. 
In this work, the reference (straight) configuration of the rod is defined as an undeformed configuration where all bulk and cross-sectional strains vanish. 
It is also intended that the natural (stress-free) configuration coincides with the reference configuration, ensuring the strain energy function is minimized, thereby causing stress to vanish. 
Consequently, the straight reference configuration aligns with both the undeformed and stress-free configurations, resulting in vanishing strains and stresses.
The tractions $N(s), H(s), M(s) \in \mathbb{R}$ are required to monotonically increase. 
This implies that any increase in bulk strains $\nu(s)$, $\eta(s)$, and $\mu(s)$ induces an increase in the corresponding tractions $N(s)$, $H(s)$, and $M(s)$, such that each traction reaches its maximum at the maximum of the respective bulk strain.
The following constitutive equations are proposed here for these tractions based on the necessary conditions required for the strain energy function $\psi$
\begin{align}
%&N, H, M:  [0,  L]  \longmapsto \mathbb{R} \nonumber \\[5pt]
&N(s)= \frac{\partial \psi}{\partial \nu}=  2 C_1 (\nu(s) -1), \nonumber \\[5pt]
&H(s)= \frac{\partial \psi}{\partial \eta}= 2 C_2 \eta(s),   \nonumber \\[5pt] 
&M(s)= \frac{\partial \psi}{\partial \mu}= 2 C_3 \mu(s),
\label{Eq_BulkForceNHMConstitutive}
\end{align}
where $C_1= C_3= E \in \mathbb{R}^+$ and $C_2= G \in \mathbb{R}^+$.  
The Young modulus $E$ and for an isotropic material the shear modulus $G$ are related by $G= \frac{E}{2 (1+ \nu_p)}$, where $\nu_p \in [0, 0.5]$ is the Poisson's ratio. 
It is notable that since $N(s)$, $H(s)$ and $M(s)$ in \eqref{Eq_BulkForceNHMConstitutive} monotonically increase, they are injective and invertible functions. 
This implies the existence of conjugate constitutive fields $\nu(N,H,M,s)$, $\eta(N,H,M,s)$ and $\mu(N,H,M,s)$. 
%
%%%%%%%%%%%% Equations of Equilibrium %%%%%%%%%%%%%%%
\section{Governing Equations for Bulk Strains }
\label{Sec_GoverningEquationsforBulkStrains}
To derive the governing equations for the system, the translational and rotational equations of equilibrium are utilized in their local forms. 
This results in three scalar-valued governing equations for the bulk strains $\nu$, $\eta$, and $\mu$. 
These equations are formulated as a system of three nonlinear ordinary differential equations (ODEs). 
The vector form of the governing equations is expressed as follows:
\begin{align}
\mathbf{f}^c_s (s) + \mathbf{f}^{cd} + \mathbf{f}^{ex} (s) + \mathbf{w}(s) = \mathbf{0},
\label{Eq_EquilibriumEqLocalTranslation}
\end{align}
\begin{align}
\mathbf{m}^c_s (s) + \left( \mathbf{r} (s) \times \mathbf{f}^c (s) \right)_s + \mathbf{m}^{cd}_s +\mathbf{m}^{ex} (s) + \mathbf{m}^w_s = \mathbf{0},
\label{Eq_EquilibriumEqLocalRotation}
\end{align}
where $\mathbf{f}^{ex} \doteq \text{d} \mathbf{F}_{vis} + \text{d} \mathbf{F}_t^{V} + \text{d} \mathbf{F}_n^{V} + \text{d} \mathbf{F}_n^{D} $ using \eqref{Eq_DistributedTCAMsForcePerLength}, \eqref{Eq_FrictionForce_dFvis}, \eqref{Eq_ForceLeftTangentRod}, \eqref{Eq_ForceLeftNormalRod},  \eqref{Eq_ForceRightNormalRod} and \eqref{Eq_BulkContactForceFcConstitutive}. 
Here, $\mathbf{w}$ is the weight of the arm per unit length of straight centerline, which is obtained from \eqref{Eq_WeightSegmentRod}.
For \eqref{Eq_EquilibriumEqLocalRotation} use \eqref{Eq_DistributedTCAMsMomentIntegral} and \eqref{Eq_BulkContactMomentMcConstitutive}, where $\mathbf{m}^{ex} \doteq \mathbf{x} \times \mathbf{f}^{ex} $ and $\mathbf{m}^w $ denotes the moment of the arm's weight in \eqref{Eq_WeightSegmentRod} about the origin of the fixed global Cartesian frame. 
%
%%%%%%%%%%%% local volume Preservation Constraint %%%%%%%%%%%%%%%
\section{Local Volume Preservation Constraint }
\label{Sec_LocallyVolumePreservationConstraint}
The constraint of local volume preservation aligns with the biological characteristics of muscular hydrostats such as octopus arms. 
In these structures, tissue incompressibility directly influences the stiffness of the hydrostat. 
Therefore, the deformation of the octopus arm is constrained by: 
\begin{align}
\frac{\partial v(s)}{\partial V(s)} = \det( F(s))= 1,
\label{Eq_LocalVolumePreservationDefinition}
 \end{align}    
where $V$ and $v$ are respectively the volumes of undeformed and deformed robot arms and $F(s) \in \mathbb{R}^{2 \times 2 }$ denotes the deformation gradient tensor:   
%Here $\mathbb{R}^{2 \times 2} $ denotes the set of all $2 \times 2 $ matrices with real entries.
%
\begin{align}
 %&F :  [0, L]  \longmapsto \mathbb{R}^{2 \times 2} \nonumber \\[5pt]
 & F(s)  \doteq \frac{\partial \mathbf{x} (s)}{\partial \mathbf{X} (s)}=  \frac{\partial \mathbf{x} (s)}{\partial X_\alpha (s)}  \otimes \mathbf{e}_\alpha,     
\label{Eq_BulkDeformationGradient_Defin}
\end{align}
where $\mathbf{X}(s)= s \, \mathbf{e}_1 + X_2(s) \mathbf{e}_2$ describes the straight arm, $X_1 \equiv s $ and $X_2 \equiv X_2(s)$ for $\alpha \in \{ 1,2\}$ and $\otimes$ denotes the tensor product of two vectors. 
The local volume preservation constraint in \eqref{Eq_LocalVolumePreservationDefinition} introduces a governing equation for the new cross-sectional normal strain $\beta \in \mathbb{R}$: 
%
%\begin{align}
%& \beta:  [0,  L]  \longmapsto \mathbb{R} \nonumber \\[5pt]
%& \beta(s) \left\{ \nu(s) - \beta(s) X_2(s) \mu(s) \right\}= 1
%\label{Eq_BetaLocallyVolumePreservationConstraint}
%\end{align}
%
%or equivalently: 
%
\begin{align}
\beta_s(s)= \frac{ - \beta ( \nu_s - \beta \partial_s X_2 \mu - \beta X_2 \mu_s) }{ \nu - 2 \beta X_2 \mu }.
\label{Eq_Beta_s_LocallyVolumePreservationConstraint}
\end{align}  
Note that the local volume preservation constraint in \eqref{Eq_LocalVolumePreservationDefinition} is satisfied only by the strain $\beta$ in \eqref{Eq_Beta_s_LocallyVolumePreservationConstraint}, without any corresponding internal force or moment associated with $\beta$.
This contrasts with scenarios where the local volume preservation constraint is enforced on the strain energy function using a Lagrange multiplier, such as pressure (stress). 
In that case, there is no strain directly conjugate to the pressure, meaning the pressure is not computed through a constitutive equation.  
%
%%%%%%%%%%%%%%%% Results and Discussions %%%%%%%%%%%%%%%%%%%
\section{Results and Discussions}
\label{Sec_ResultsAndDiscussions}
In this section, we apply the previously discussed general formulations to a specific case study. 
This case study focuses on an octopus-like soft robot arm with a straight undeformed configuration and a truncated conic geometry that is characterized by a radius of $\left \vert X_2(s) \right \vert$ \cite{Renda2014_a}
\begin{align}
%&X_2: [0, L]  \longmapsto \mathbb{R} \nonumber \\[5pt]
&\left \vert X_2(s)  \right \vert= \left ( \frac{R_{min} - R_{max}}{L}  \right) s + R_{max}
\label{Eq_RodRadius}
\end{align}  
such that $R_{max} \geq R_{min} > 0$ denote the maximum and minimum radii of the undeformed rod cross sections, respectively. 
The distance $y_{ij}^{c}(s) \in \mathbb{R}$ of the TCAMs $j \in \{ 1, 2\}$ in segment $i \in \mathbb{Z}^+$ from the undeformed centerline is calculated as:
\begin{align}
% & y_{ij}^{c}: [0,  L_i]  \longmapsto \mathbb{R} \nonumber \\[5pt]
 & y_{ij}^{c}(s) = (-1)^{(j-1)} \left \{ \left( \frac{b_i - a_i}{L_i} \right) s + a_i \right\}, 
\label{Eq_TCAMsDistance}
\end{align} 
where $0 < b_i < a_i $ such that $a_i \in \mathbb{R}^+ $ and $b_i \in \mathbb{R}^+ $ represent the distances of each TCAM in segment $i \in \mathbb{Z}^+$ from the undeformed centerline of the rod successively at the beginning and end of that segment.
In this setup, electro-thermo carbon fibers/silicone rubber TCAMs are employed to actuate the octopus robot arm. 
These TCAMs are driven by a sinusoidal input voltage \( v_{vol}(t) = V_{amp} \sin(\omega_{vol} t) \), where \( V_{amp} = 9 \, \text{V} \) is the amplitude and \( \omega_{vol} = 1 \) rad/sec is the angular frequency. 
Table \ref{Table_TCAMs_Parameters} tabulates the parameters of these TCAMs \cite{Hammond2022a}.
%
%%%%%%%%%%%%%%%% Table %%%%%%%%%%%%%%%%%%%
\begin{table}[htbp]
  \caption{Parameters of TCAMs.}  
  \label{Table_TCAMs_Parameters}
  \centering
  \begin{tabular}{cp{2 cm}cp{2 cm}}
  \toprule
\textbf{Parameters} & \textbf{Values} \\[3pt]
\midrule
  $ m $  &  $ 0.106  $   Kg \\[3pt]
 $ \bar{m} $  &  $ 0.106 $   Kg \\[3pt]
 $\bar{S} $  &  $ 0.46 $   m \\[3pt]
 $ \mathcal{\bar{L}}_{amb}$  &  $ 0.418 $   m \\[3pt]
 $ n$  &  $ 200 $ \\[3pt]
 $ r^o_{tm}$  &  $ 3.6 \times 10^{-4}$  m \\[3pt]
 $ T_{amb} $  &  $ 23 $  $\, ^\circ\text{C}$ \\[3pt]
 $ CTE$  &  $ 3 \times 10^{-4}$  $\text{C}^{-1} $ \\[3pt]
 $ C_t$  &  $0.162 $  $ \text{JC}^{-1}$ \\[3pt]
 $R_{vol} $  &  $18 $  $ \Omega$ \\[3pt]
 $ E_x$  &  $4.67 \times 10^{8} $  Pa \\[3pt]
 $ G_{yz}$  &  $2.2 \times 10^{7} $  Pa \\[3pt]
 $ \lambda$  &  $ 0.0086$  $\text{WC}^{-1} $ \\[3pt]
 \bottomrule
 \end{tabular}
\end{table}
%
%%%%%%%%%%%%%%%%%%%%%%%%%%%%%%%%%%%%%%%
\subsubsection{Results}
\label{Sec_Results}
\begin{figure}[t]
\centering
\begin{minipage}{0.48\textwidth}
\centering
\begin{subfigure}[t]{\textwidth}
\centering
\includegraphics[width=\textwidth, height=1.8in]{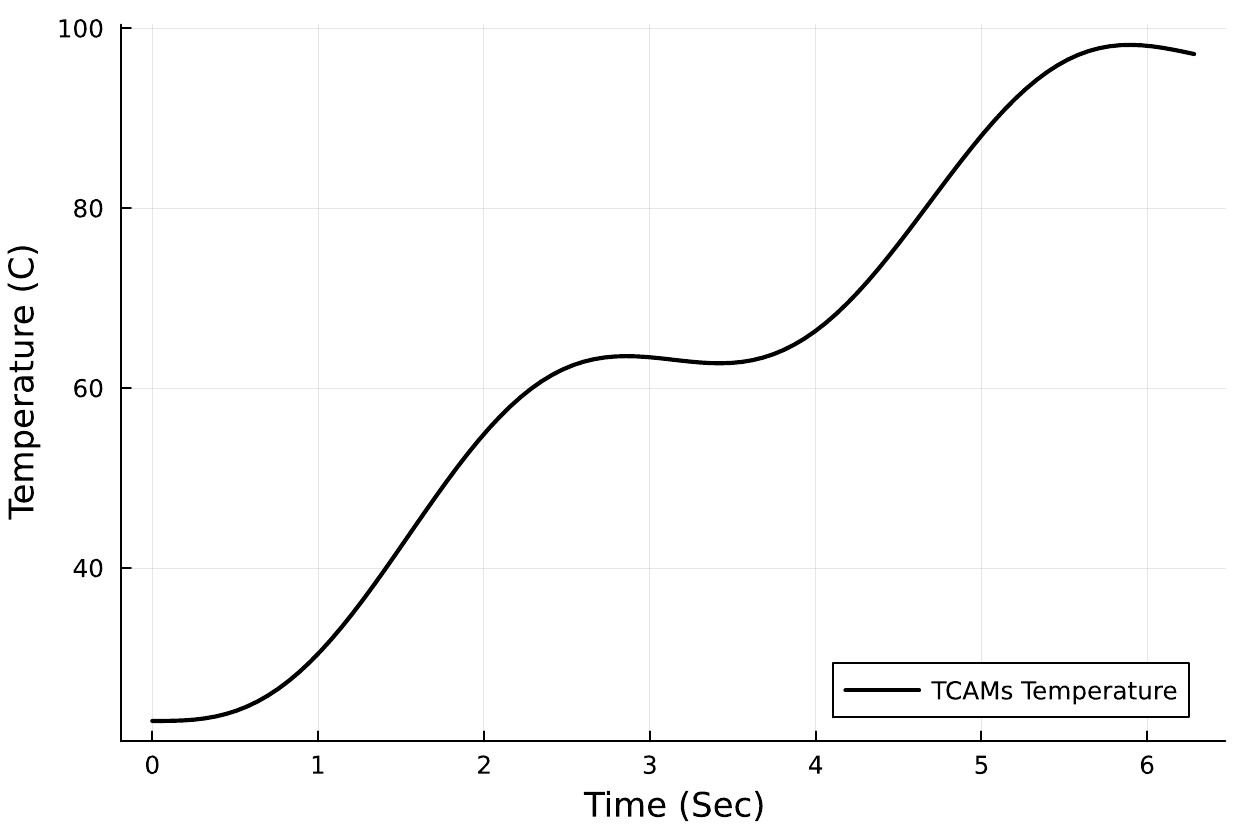}
\captionsetup{justification=centering}
\caption{TCAMs temperature.}
\label{Fig_Temp_TCAMs_Time}
\end{subfigure}
\end{minipage}
\hspace{0.04\textwidth}
\begin{minipage}{0.48\textwidth}
\centering
\begin{subfigure}[t]{\textwidth}
\centering
\includegraphics[width=\textwidth, height=1.8in]{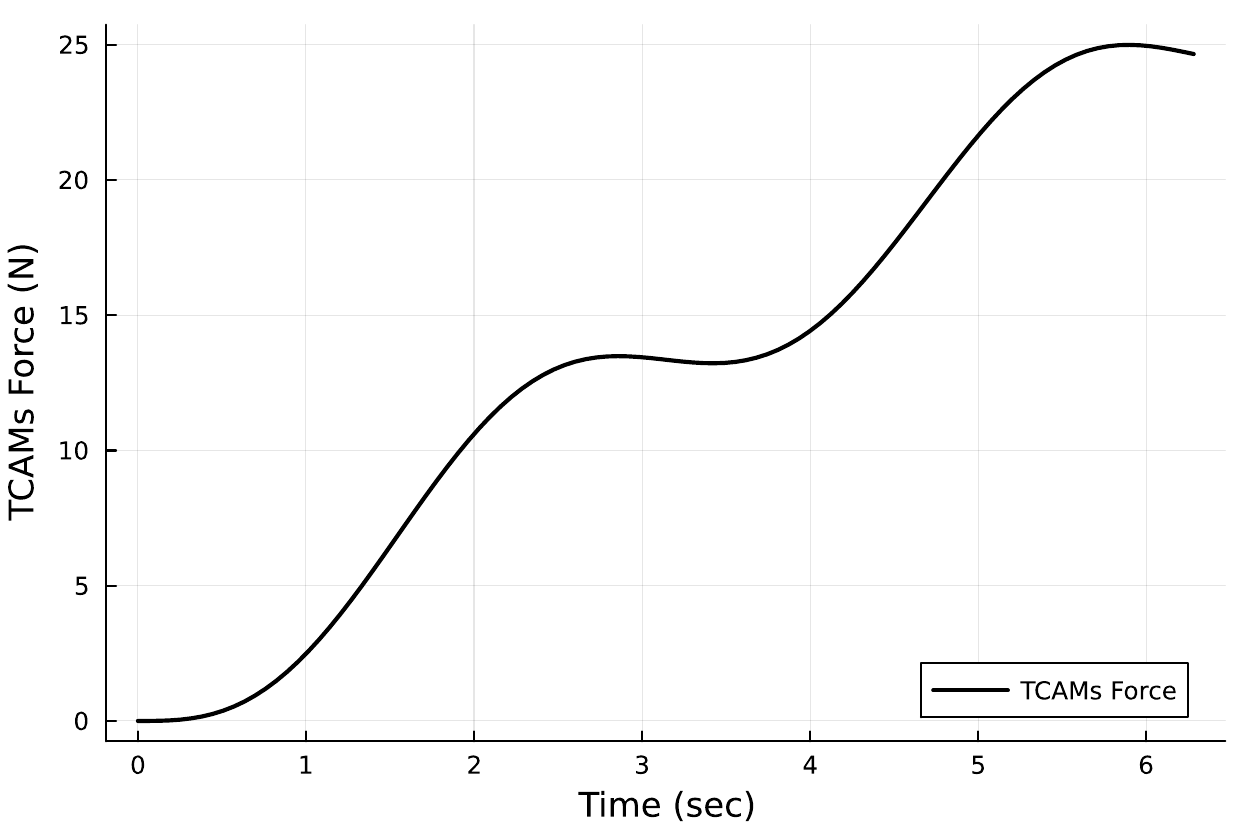}
\captionsetup{justification=centering}
\caption{TCAMs tensile force.}
\label{Fig_F_TCAMs_Time}
\end{subfigure}
\end{minipage}
\caption{TCAMs temperature and tensile force, actuated by a sinusoidal voltage.}
\label{Fig_TempForce_TCAMs_Time}
\end{figure}

\begin{figure*}%[H]
\centering
\begin{minipage}{0.48\textwidth}
\centering
\begin{subfigure}[t]{1.0\textwidth}
\centering
\includegraphics[height=2.0in]{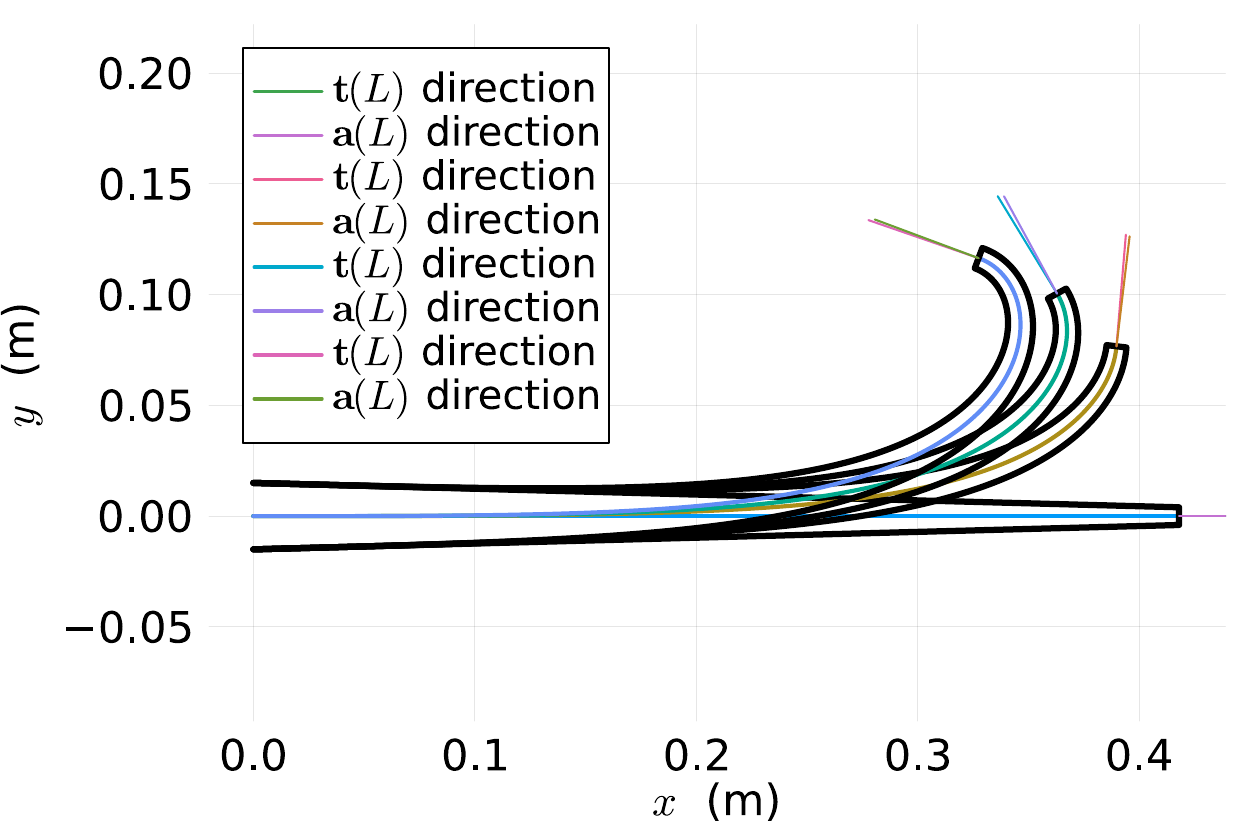}
\captionsetup{justification=centering}
\caption{Deformations for some $T_{11} \in [0, 20]$ and $v_{\infty}= 0.2$ m/s.}
\label{Fig_Plot_Arm_Some}
\end{subfigure}
\end{minipage} 
\begin{minipage}{0.48\textwidth}
\centering
\begin{subfigure}[t]{1.0\textwidth}
\centering
\includegraphics[height=1.3in]{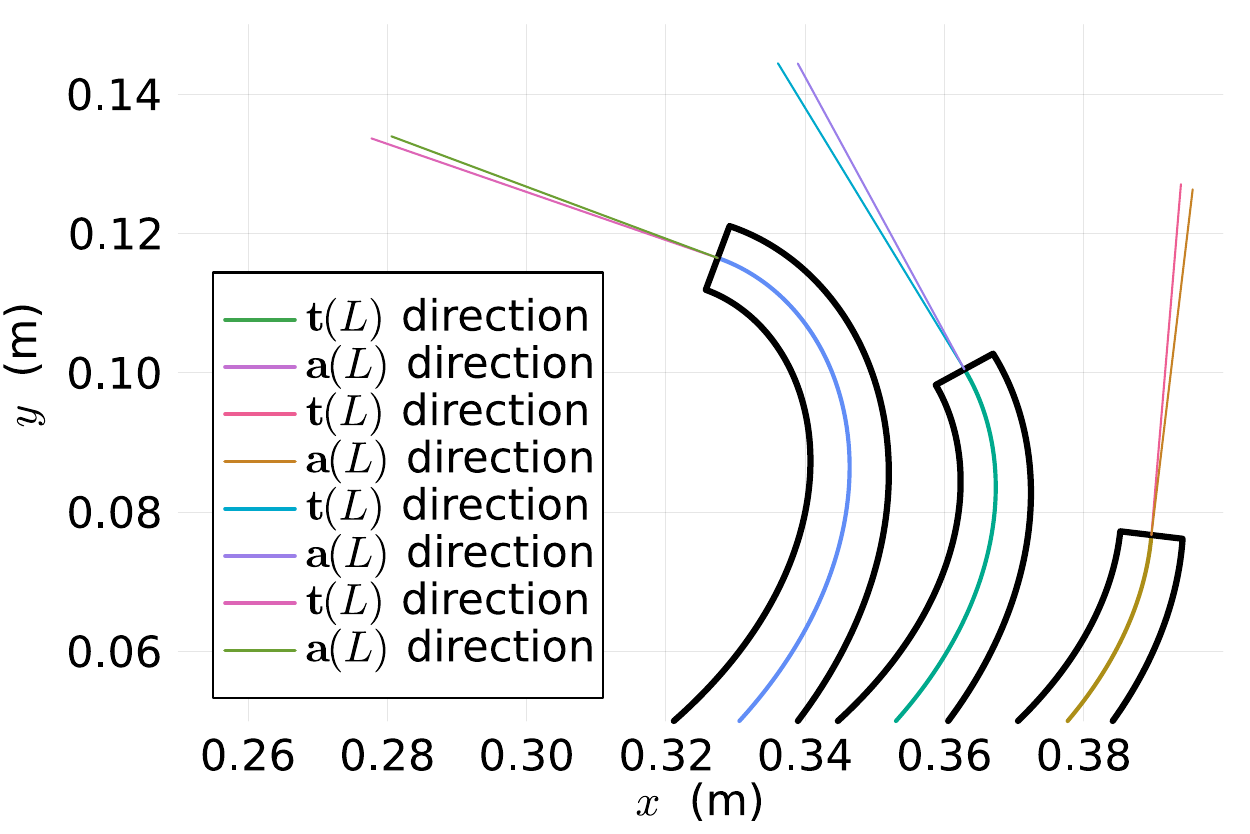}
\captionsetup{justification=centering}
\vspace{1.1in}
\label{Fig_Plot_Arm_Some_Zoom}
\end{subfigure}
\end{minipage}   
\vspace{0.05in}

\centering
\begin{minipage}{0.48\textwidth}
\centering
\begin{subfigure}[t]{1.0\textwidth}
\centering
\includegraphics[height=2.1in]{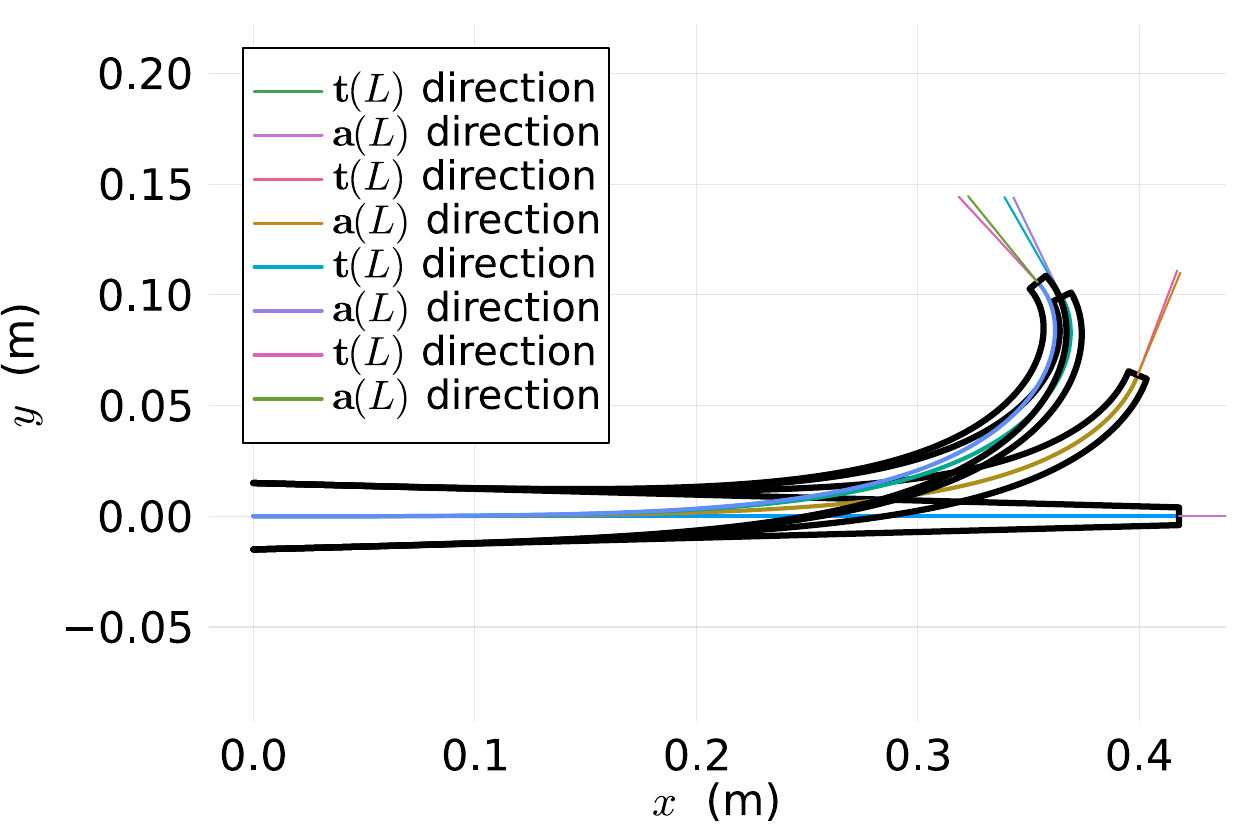}
\captionsetup{justification=centering}
\caption{Deformations for some $T_{11} \in [0, 20]$ and $v_{\infty}= 0.4$ m/s.}
\label{Fig_Plot_Arm_Some2}
\end{subfigure}
\end{minipage} 
\begin{minipage}{0.48\textwidth}
\centering
\begin{subfigure}[t]{1.0\textwidth}
\centering
\includegraphics[height=1.4in]{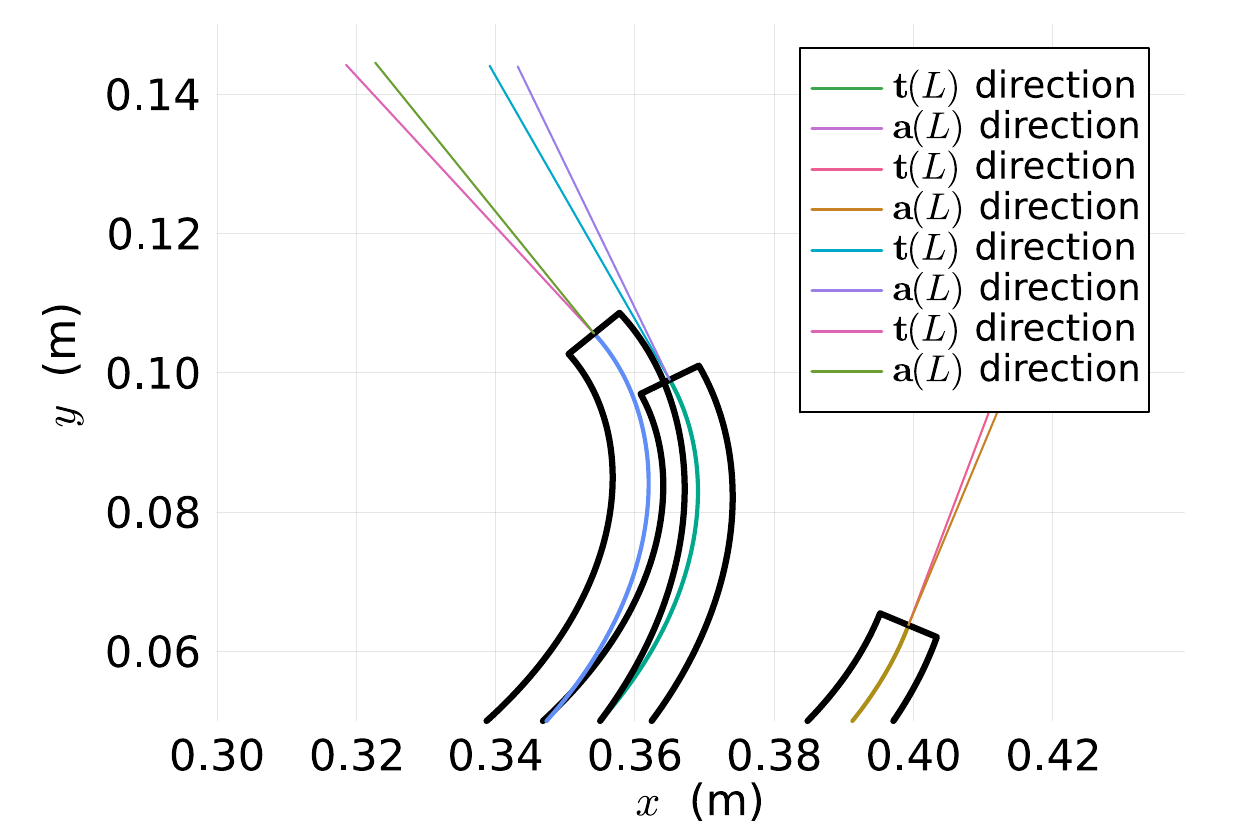}
\captionsetup{justification=centering}
\vspace{1.1in}
\label{Fig_Plot_Arm_Some2_Zoom}
\end{subfigure}
\end{minipage}   
\vspace{0.1in}

\centering
\begin{subfigure}[t]{1.0\textwidth}
\centering
\includegraphics[height=2.1in]{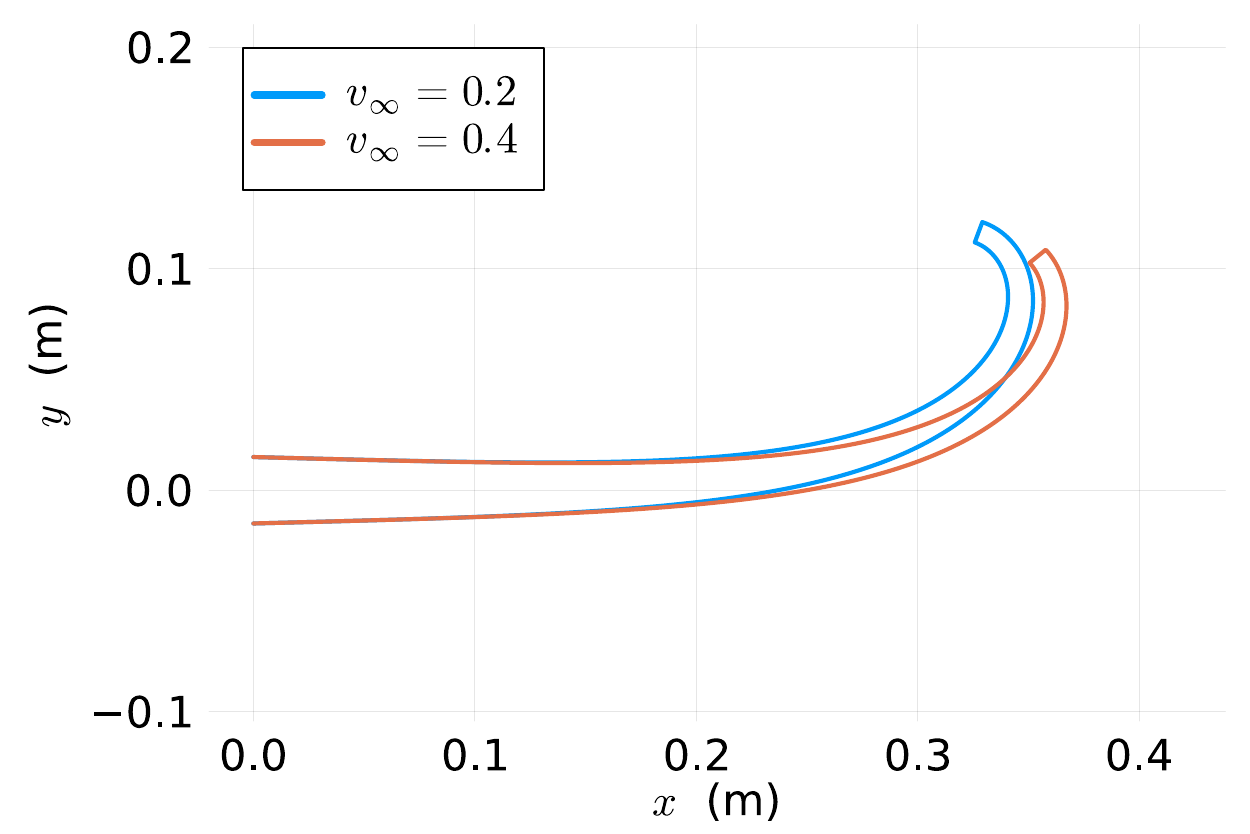}
\captionsetup{justification=centering}
\caption{Deformations for $T_{11}= 20$ N and $v_{\infty}= 0.2, \, 0.4$ m/s.}
\label{Fig_Plot_Arm_VInf}
\end{subfigure}

\caption{Deformations for various tensions of TCAMs $T_{11} \in [0, 20]$ N and water free-stream velocities $v_{\infty}= 0.2, \, 0.4$ m/s.}
\label{Fig_PlotArm12_T11}
\end{figure*}

Fig.~\ref{Fig_Temp_TCAMs_Time} shows that the temperature of the electro-thermo TCAMs, stimulated by the sinusoidal voltage, rises to $98 , ^\circ\text{C}$. 
Correspondingly, the TCAMs' tensile force in Fig.~\ref{Fig_F_TCAMs_Time} increases similarly, reaching $25$ N. 
It is important to note that the functionality of the TCAMs is highly dependent on temperature, with potential malfunctions occurring beyond a certain threshold. 
Fig.~\ref{Fig_Temp_TCAMs_Time} indicates that the TCAMs are operating within a feasible temperature range.
%
%%%%%%%%%%%%% Figure %%%%%%%%%%%%%%%%%%%%%
%
%%%%%%%%%%%%%%%%%%%%%%%%%%%%%%%%%%%%%%
The system of four coupled nonlinear ODEs \eqref{Eq_EquilibriumEqLocalTranslation}, \eqref{Eq_EquilibriumEqLocalRotation} and \eqref{Eq_Beta_s_LocallyVolumePreservationConstraint} forms a quasi-static model. 
This model is used to study the effects of hydrostatic and dynamic forces from steady-state fluid flow on the bending motion of an octopus-like robot arm driven by TCAMs. 
The deformed configuration of the arm is analyzed statically at each incremental increase in the tension of the TCAMs. 
The model is numerically solved for the four unknown strain fields: normal strain $\nu$, shear strain $\eta$, bending strain $\mu$, and cross-sectional normal strain $\beta$, under the boundary conditions (BCs) for the fixed support arm: $\nu_o= \beta_o= 1$ and $\eta_o= \mu_o= \beta '_o= 0$.
In this section, the application of the model to two case studies involving free-stream velocities $v_\infty= 0.2$ m/s and $0.4$ m/s of steady-state water flow is presented. 
For each case study, a step size of $0.5$ N is used to incrementally increase the tension $T_{11} \in [0, 20]$ N of the TCAMs, allowing for a static analysis of the arm's deformation under water at each tension level. 
The parameters and dimensions for both case studies are provided in Table~\ref{Table_RodWaterParameters}.
%
%%%%%%%%%%%%%%%% Table %%%%%%%%%%%%%%%%%%%
\begin{table}[htbp]
  \caption{Parameters of silicone rubber and water.}  \label{Table_RodWaterParameters}
  \centering
  \begin{tabular}{cp{2cm}cp{2cm}}
  \toprule
\textbf{ Parameters }	& \textbf{ Values }  \\[3pt]
\midrule
   $ E $  &  $ 10 $  MPa     \\[3pt]
 $ \nu_p $  & $ 0.5 $   \\[3pt]
  $\rho $ & $ 1.1 $ $\text{Kg}/ \text{dm}^3 $   \\[3pt]
  $g^r$  & $9.81$ $\text{m} / \text{s}^2$ \\[3pt]
  $R_{min}$ & $4$  mm  \\[3pt]
  $ R_{max}$ & $15$  mm \\[3pt]
 $L$   & $418$ mm  \\[3pt]
 $a_1$  & $12$ mm  \\[3pt]
  $b_1$ & $1$  mm  \\[3pt]
 $\rho_w$  & $0.998$ $\text{Kg}/ \text{dm}^3 $   \\[3pt]
$\mu_v$ & $1.002$ \text{mPa.s} \\[3pt]    
$v_\infty$ & $0.2$, $0.4$  m/s \\[3pt]
%$h_\infty$ & $2$, $3$ cm \\[3pt]
%$y$ & $2$, $3$ mm \\[3pt]
$T_{11}$ & [0, 20] N \\[3pt]
$T_{12}$ & 0.0 N \\[3pt]
\bottomrule
\end{tabular}
\end{table}
%
%%%%%%%%%%%%%%%%%%%%%%%%%%%%%%%%%%%%%%%
Fig.~\ref{Fig_PlotArm12_T11} illustrates the deformations of the arm for various tensions $T_{11} \in [0, 20]$ N applied by the TCAMs and for two water free-stream velocities, $v_{\infty} = 0.2$ m/s and $0.4$ m/s.
The direction of the tangent vector $\mathbf{t}(L)$ in \eqref{Eq_SpatialDerivative_rVector} to the centerline of the arm at $s=L$, and the direction of the unit director $\mathbf{a}(L)$ at $s=L$, are displayed on each configuration of the rod in both Fig.~\ref{Fig_Plot_Arm_Some} and \ref{Fig_Plot_Arm_Some2}.
It is notable that traditional rod theories do not account for any deformation or rigid rotation of the cross section. 
Consequently, the cross section at any $s \in [0, L]$ always remains perpendicular to the tangent line of the centerline. 
In other words, the tangent vector $\mathbf{t}(L)$ in Fig.~\ref{Fig_PlotArm12_T11} signifies the orientation of the cross section of the rod at $s= L$, before any rigid rotation. 
On the other hand, in the standard Cosserat theory of rods, the unit director $\mathbf{a}(s)$ is defined such that it always remains perpendicular to the cross section of the rod through deformation and signifies the orientation of the cross section after rigid rotation. Therefore, each unit director $\mathbf{a}(L)$ in Fig.~\ref{Fig_PlotArm12_T11} shows the orientation of the rotated cross section at $s= L$ for the corresponding configuration.  \par
Fig.~\ref{Fig_PlotArm12_T11} demonstrates that as the tension of the TCAMs $ T_{11} \in [0, 20] $ increases, the overall deformation of the arm gradually amplifies. 
This is observed in both scenarios of water free-stream velocities: $ v_{\infty}= 0.2 $  $m/s $ (Fig.~\ref{Fig_Plot_Arm_Some}) and $ v_{\infty}= 0.4 $  $m/s$ (Fig.~\ref{Fig_Plot_Arm_Some2}). 
Specifically, at $ s = L $, the tip of the arm reaches heights of $ y = 11.65 $ $cm$ and $ 10.56 $ $cm$ respectively for these velocities under the constant tension $ T_{11} = 20 $ $N$.
This implies that for the same tension $ T_{11} $, the overall deformation of the arm, measured by the height of its tip at $ s = L $, is less pronounced at higher water free-stream velocities. 
This reduction in bulk deformation is primarily attributed to the effects of drag force and the dynamic forces exerted by the water flow on the arm's surface, where both the drag force and resulting dynamic force are proportional to the square of the flow velocity.
The dynamic force in a steady-state flow results from changes in linear momentum (velocity vector) within the fluid's control volume between two open surfaces. 
This alteration in momentum does not solely stem from friction between the arm's surface and the water, as it can occur even in the flow of an inviscid fluid over a frictionless body surface.
Fig.~\ref{Fig_Plot_Arm_VInf} visually compares the bulk deformation of arm configurations under the constant tension of TCAMs $ T_{11} = 20 $ $N$ for two scenarios: water free-stream velocities $ v_{\infty} = 0.2 $ $m/s$ and $ v_{\infty} = 0.4 $ $m/s$. 
This comparison underscores that, when subjected to the same tension $ T_{11} $, the arm experiences less deformation at higher free-stream flow velocities.  \par 
%
%%%%%%%%%%%%%%%%% Figure %%%%%%%%%%%%%%%%%%
\begin{figure*}[t]  % Use figure* to span across both columns
\centering
\begin{minipage}{0.45\textwidth}  % Adjust the width as needed for your layout
\centering
\begin{subfigure}[t]{\textwidth}
\centering
\includegraphics[height=1.5in]{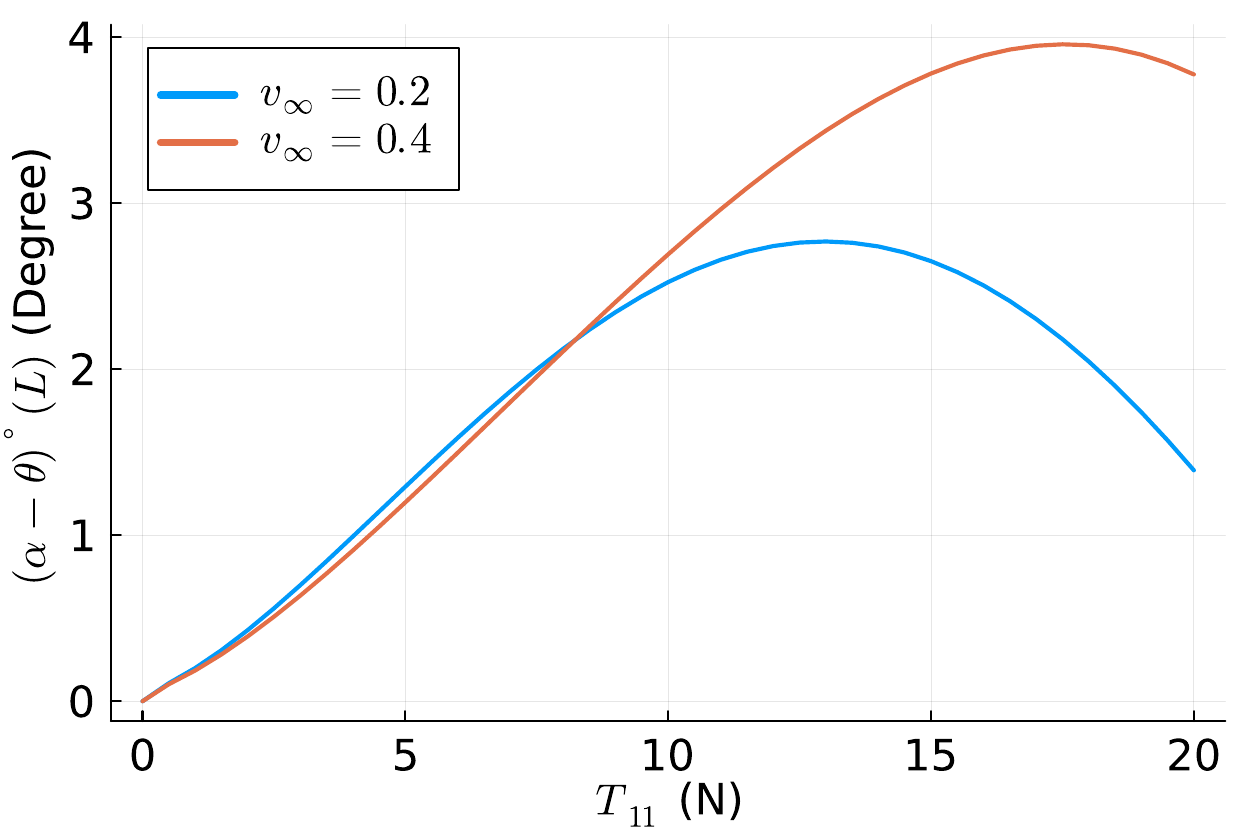}
\caption{Tilt angle $(\alpha - \theta)^\circ \in [0, 180]$ at $s= L$.}
\label{Fig_PlotAll_AlphaThetaL_DEGREE}
\end{subfigure}
\end{minipage}
\hspace{0.5cm}  % Adjust spacing between minipages
\begin{minipage}{0.45\textwidth}  % Adjust the width as needed for your layout
\centering
\begin{subfigure}[t]{\textwidth}
\centering
\includegraphics[height=1.5in]{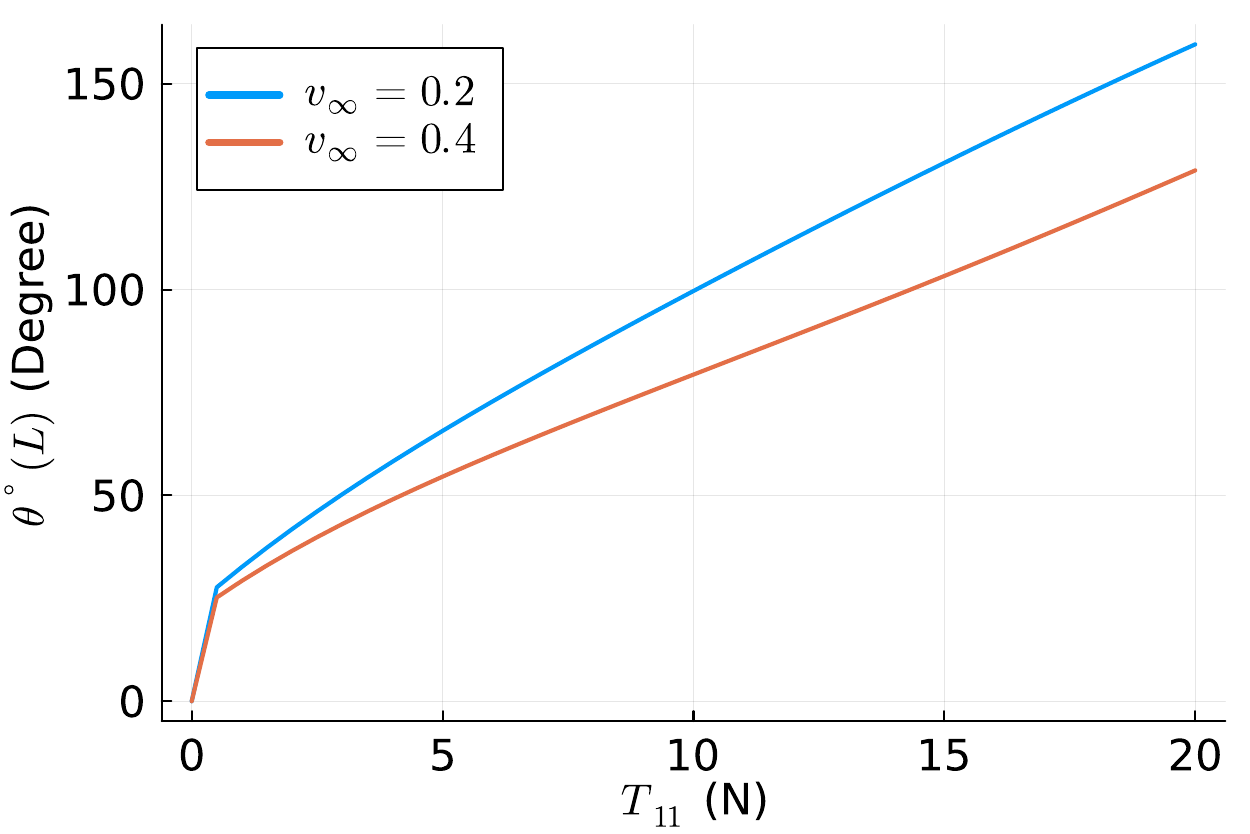}
\caption{Angle $\theta^\circ \in [0, 180]$ at $s= L$.}
\label{Fig_PlotAll_ThetaL_DEGREE}
\end{subfigure}
\end{minipage}

\vspace{0.5cm}  % Adjust vertical spacing between rows

\begin{minipage}{0.45\textwidth}  % Adjust the width as needed for your layout
\centering
\begin{subfigure}[t]{\textwidth}
\centering
\includegraphics[height=1.5in]{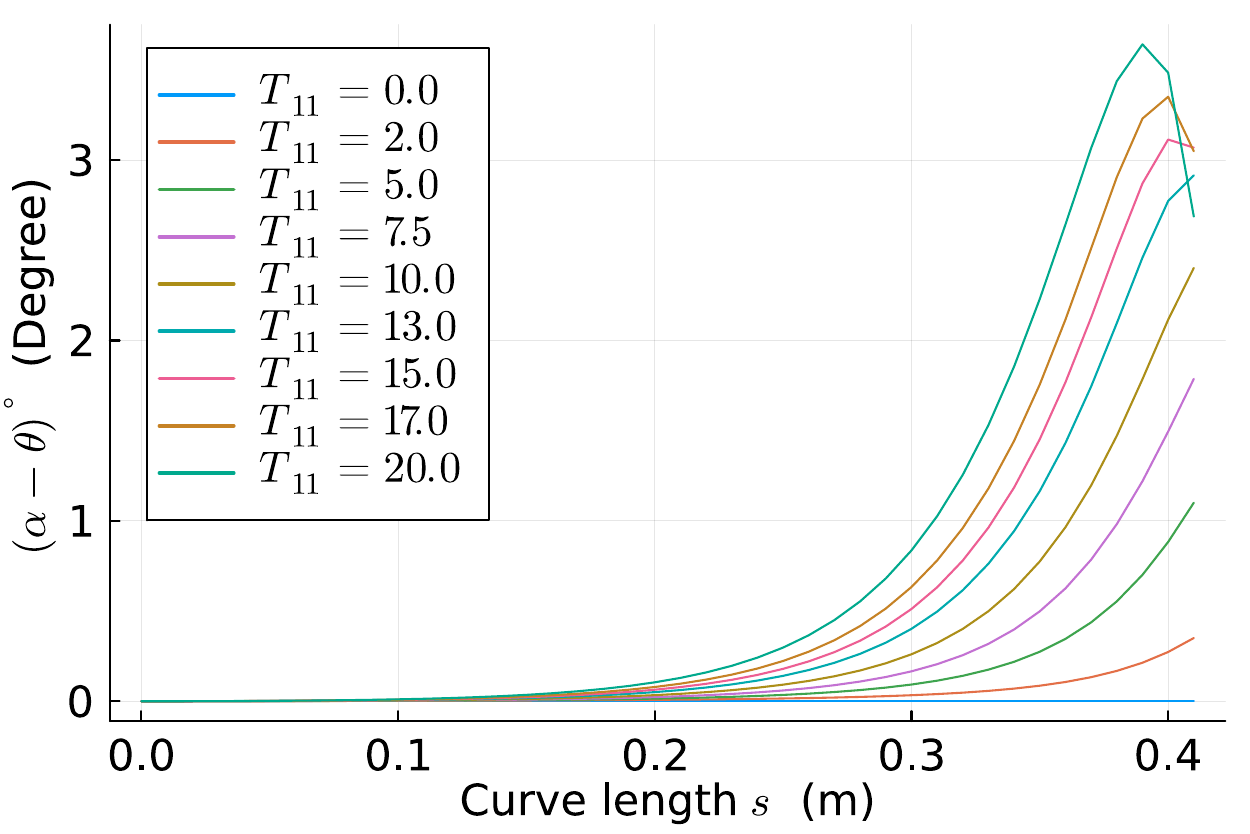}
\caption{Tilt angle $(\alpha - \theta)^\circ \in [0, 180]$ along the rod for $v_{\infty}= 0.2$ m/s.}
\label{Fig_PlotSome_AlphaTheta_DEGREE}
\end{subfigure}
\end{minipage}
\hspace{0.5cm}  % Adjust spacing between minipages
\begin{minipage}{0.45\textwidth}  % Adjust the width as needed for your layout
\centering
\begin{subfigure}[t]{\textwidth}
\centering
\includegraphics[height=1.8in]{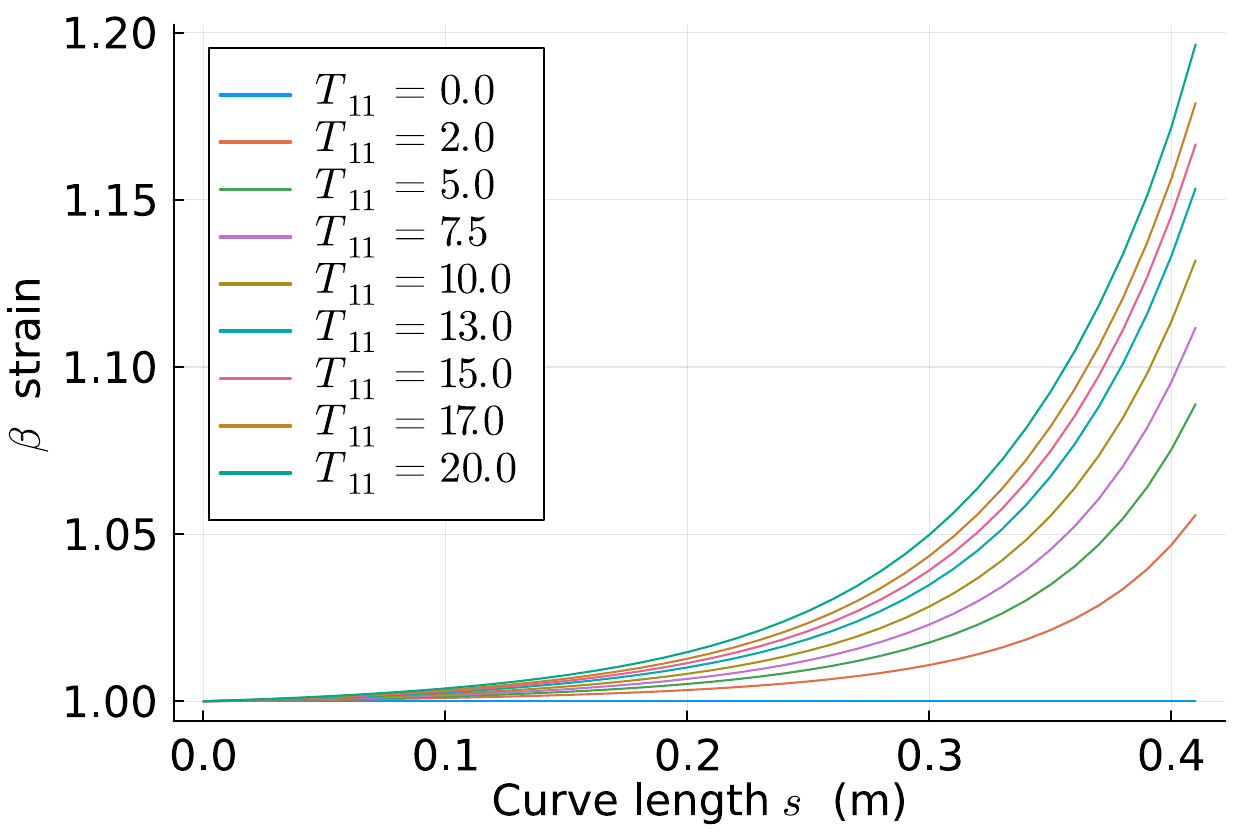}
\caption{Normal strain $\beta(s)$ along the rod for $v_{\infty}= 0.2$ m/s.}
\label{Fig_PlotSome_Beta}
\end{subfigure}
\end{minipage}

\caption{Variations of angles $(\alpha - \theta)^\circ$, $\theta^\circ$, and cross-sectional strain $\beta(s)$.}
\label{Fig_AlphaTheta_Alpha_AlphaTheta_Beta}
\end{figure*}

%%%%%%%%%%%%%%%%%%%%%%%%%%%%%%%%%%%%%%
Fig.~\ref{Fig_PlotAll_AlphaThetaL_DEGREE} illustrates the variation of the \textit{tilt angle}, $ (\alpha - \theta)(s) \in [0, 180] $, of the rod's cross section at $ s = L $ (see Fig.~\ref{Fig_DefArm}) across tensions of TCAMs $ T_{11} \in [0, 20] $ $N$, for water free-stream velocities $ v_{\infty} = 0.2 $ and $ v_{\infty} = 0.4 $ $m/s$.
From Fig.~\ref{Fig_DefArm}, $\alpha(s) \in \mathbb{R}$ and $\theta(s) \in \mathbb{R}$ respectively signify the orientations of the cross section before and after the rigid rotation of the cross section.
In essence, $ (\alpha - \theta)^\circ (L) $ depicted in Figure~\ref{Fig_PlotAll_AlphaThetaL_DEGREE} indicates how much the cross section deviates from its original perpendicular orientation due to the rod's deformation.
Figure~\ref{Fig_PlotAll_AlphaThetaL_DEGREE} demonstrates that in the straight arm, the tangent vector $\mathbf{t}(L)$ aligns with the unit director $\mathbf{a}(L)$ (see Figs.~\ref{Fig_Plot_Arm_Some} and \ref{Fig_Plot_Arm_Some2}) for both water free-stream velocities.
%The tilt angle increases to a maximum of $(\alpha - \theta)^\circ (L)= 2.77^\circ$ at TCAMs tension $T_{11}= 13$ N for $v_{\infty}= 0.2$ m/s, and to $(\alpha - \theta)^\circ (L)= 3.96^\circ$ at $T_{11}= 17.5$ N for $v_{\infty}= 0.4$ m/s. 
%It subsequently decreases to $(\alpha - \theta)^\circ (L)= 1.39^\circ$ and $(\alpha - \theta)^\circ (L)= 3.77^\circ$ at TCAMs tension $T_{11}= 20$ N, respectively for $v_{\infty}= 0.2$ and $v_{\infty}= 0.4$ m/s.
Figure~\ref{Fig_PlotAll_AlphaThetaL_DEGREE} illustrates that $(\alpha - \theta)^\circ (L)$ is consistently positive for both $v_{\infty}= 0.2$ and $v_{\infty}= 0.4$ m/s across all TCAMs tensions $T_{11} \in [0, 20]$ N. 
This implies $\alpha(L) \geq \theta(L)$ algebraically, indicating a clockwise tilt of the rod's cross section at every tension and for both water velocities.
The clockwise rotation is visually confirmed by the orientations of $\mathbf{t}(L)$ and $\mathbf{a}(L)$ in Figs.~\ref{Fig_Plot_Arm_Some} and \ref{Fig_Plot_Arm_Some2}.
The orientation of the cross section at $s= L$ is influenced by two primary factors: 1) the difference in curvatures between the ventral and dorsal surfaces of the rod, and 2) the concentrated force exerted by the TCAMs $\mathbf{f}_{ij}^{cc}(L)$ as defined in \eqref{Eq_ConcentratedTCAMsForce}. 
For TCAM tensions $T_{11} \in [0, 20]$ N and $T_{12}= 0$ N, which produce counterclockwise bent configurations as seen in Figs.~\ref{Fig_Plot_Arm_Some} and \ref{Fig_Plot_Arm_Some2}, the curvature of the ventral surface at any point $s \in [0, L]$ is generally greater than or equal to the curvature of the dorsal surface at the same point $s \in [0, L]$.
The cumulative effect of these local curvature differences along the ventral and dorsal surfaces drives the clockwise rotation of the cross section at $s= L$, as depicted in Fig.~\ref{Fig_PlotArm12_T11}.
The initial increase in the tilt angle $(\alpha - \theta)^\circ (L)$ shown in Fig.~\ref{Fig_PlotAll_AlphaThetaL_DEGREE} is attributed to the heightened curvatures of the ventral and dorsal surfaces of the arm for higher TCAM tensions. %$T_{11} \in [0, 13]$ $N$ at $v_{\infty}= 0.2$ $m/s$, and $T_{11} \in [0, 17.5]$ $N$ at $v_{\infty}= 0.4$ $m/s$. 
This results in increasing differences in local curvatures, consequently leading to a larger tilt angle $(\alpha - \theta)^\circ (L)$ at $s= L$.
Conversely, the decreasing trend of the tilt angle, observed at $T_{11} \in [13, 20]$ N for $v_{\infty}= 0.2$ m/s and at $T_{11} \in [17.5, 20]$ N for $v_{\infty}= 0.4$ m/s, is driven by the increasing concentrated force of the TCAMs $\mathbf{f}_{ij}^{cc}(L)$. 
This concentrated force induces a counterclockwise rotation in the cross section at $s= L$, opposing the clockwise rotation caused by the curvature differences along the ventral and dorsal surfaces. 
%
%%%%%%%%%%%% Figure %%%%%%%%%%%%%%%%%%%%%%
%
%%%%%%%%%%%%%%%%%%%%%%%%%%%%%%%%%%%%%%
Fig.~\ref{Fig_PlotAll_AlphaThetaL_DEGREE} illustrates that the tilt angle reaches larger values for a water free-stream velocity of $v_{\infty}= 0.4$ m/s compared to $v_{\infty}= 0.2$ m/s, under the same tensions $T_{11} \in [9, 20]$. 
This accelerated increase in tilt angle can be attributed to the faster growth of angle $\theta^\circ \in [0, 180]$ observed at $v_{\infty}= 0.2$ m/s, as shown in Figure~\ref{Fig_PlotAll_ThetaL_DEGREE}.
Fig.~\ref{Fig_PlotSome_AlphaTheta_DEGREE} displays the variation of the tilt angle along the arm for selected tensions $T_{11} \in [0, 20]$ N and at $v_{\infty}= 0.2$ m/s. 
The positive tilt angle implies clockwise rotation of the cross section along the arm.  
%As discussed earlier, the orientation of the rotated cross section along the deformed rod is governed by the difference in curvatures between the ventral and dorsal surfaces. 
%In Fig.~\ref{Fig_Plot_Arm_Some}, the curvature of the ventral surface exceeds or equals that of the dorsal surface along the counterclockwise-bent arm, driving a clockwise rotation of the cross section and resulting in a positive tilt angle $(\alpha - \theta)^\circ (s)$ along the arm.
%Fig.~\ref{Fig_PlotSome_AlphaTheta_DEGREE} also shows that as TCAM tension increases, the upward trend in tilt angle along the rod reverses near the rod's end, transitioning into a downward trend towards $s= L$. 
Note that, the downward trend near the rod's end is influenced by the increasing concentrated force of TCAMs at higher tensions.  %inducing a counterclockwise rotation near the rod's end that counteracts the clockwise rotation driven by bulk deformation and curvature differences along the arm's surfaces. \par
\par
\begin{figure*}[t]  % Use figure* to span across both columns
\centering
\begin{minipage}{0.45\textwidth}  % Adjust the width as needed for your layout
\centering
\begin{subfigure}[t]{\textwidth}
\centering
\includegraphics[height=1.5in]{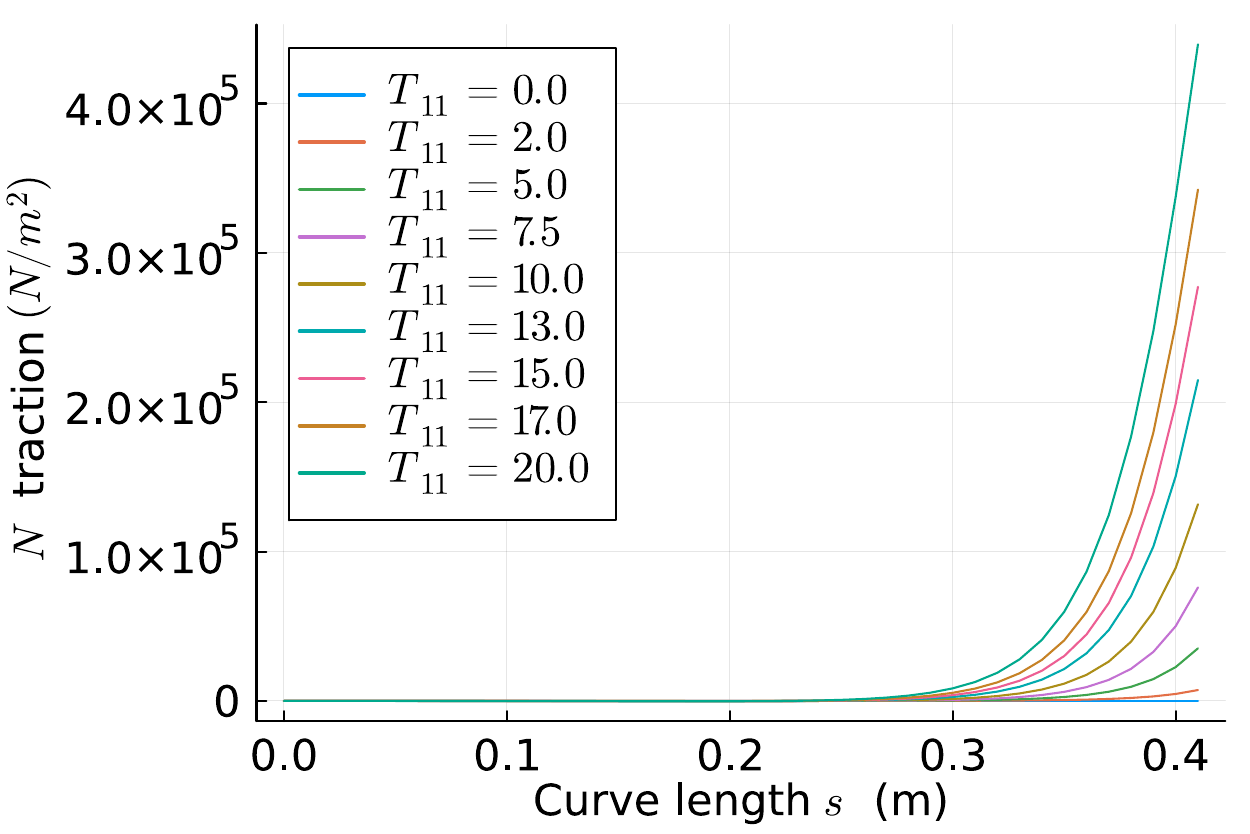}
\caption{Normal traction $N(s)$ along the rod.}
\label{Fig_PlotSome_NStress}
\end{subfigure}
\end{minipage}
\hspace{0.5cm}  % Adjust spacing between minipages
\begin{minipage}{0.45\textwidth}  % Adjust the width as needed for your layout
\centering
\begin{subfigure}[t]{\textwidth}
\centering
\includegraphics[height=1.5in]{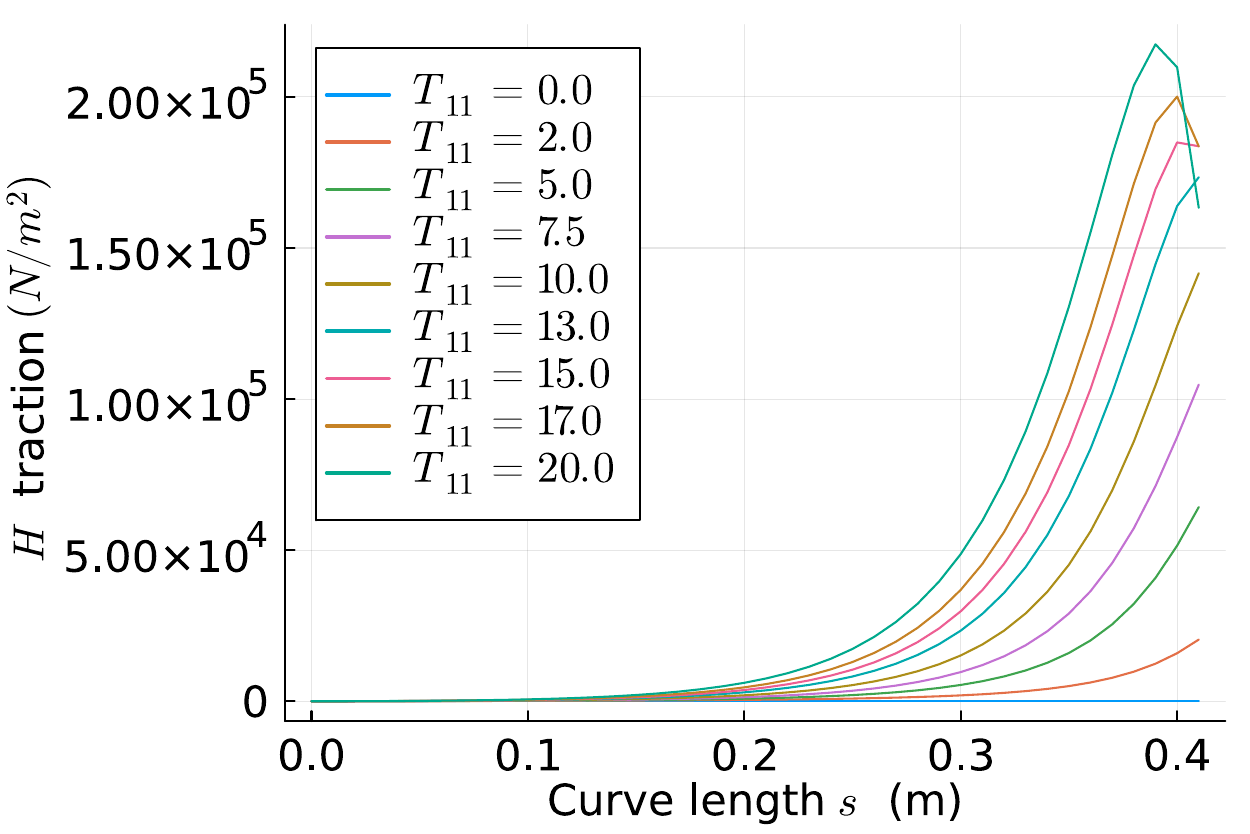}
\caption{Shear traction $H(s)$ along the rod.}
\label{Fig_PlotSome_HStress}
\end{subfigure}
\end{minipage}

\vspace{0.5cm}  % Adjust vertical spacing between rows

\begin{minipage}{0.45\textwidth}  % Adjust the width as needed for your layout
\centering
\begin{subfigure}[t]{\textwidth}
\centering
\includegraphics[height=1.5in]{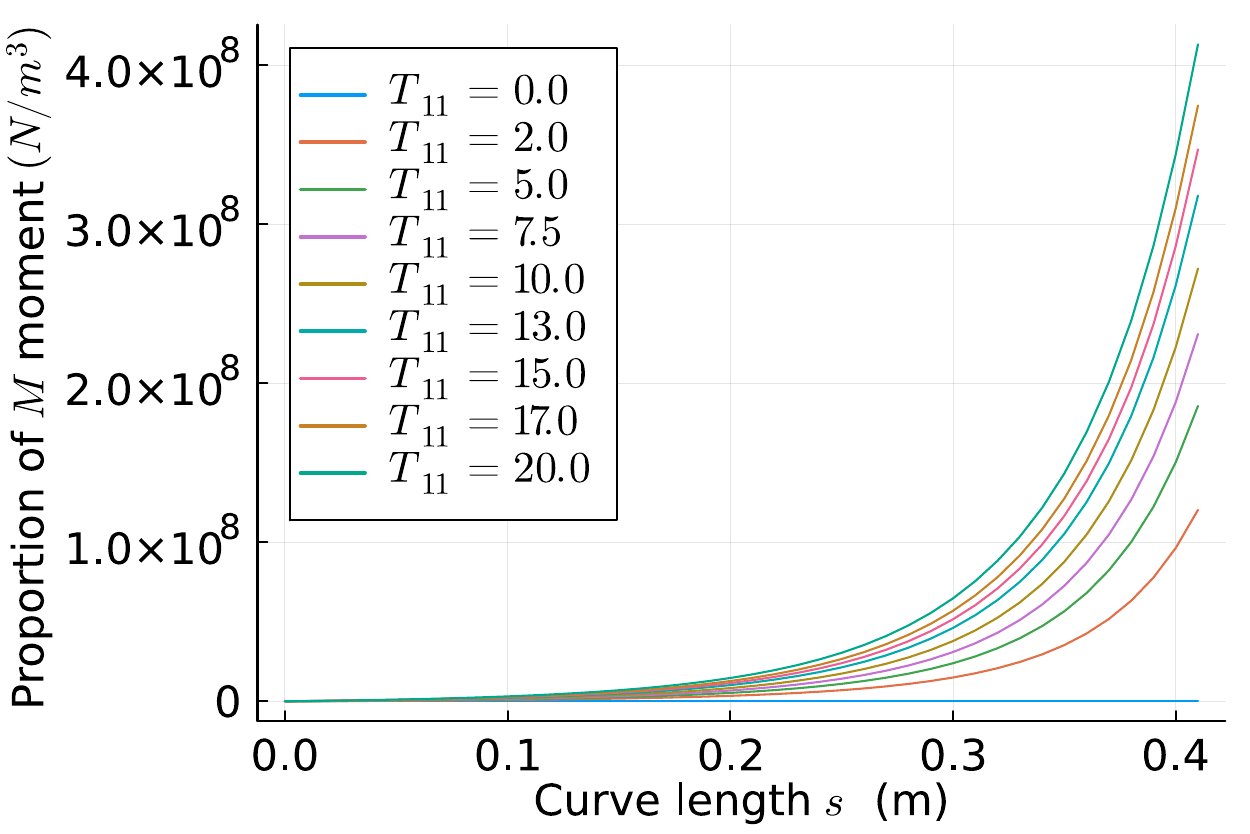}
\caption{Bending moment $M(s)$ along the rod.}
\label{Fig_PlotSome_MStress}
\end{subfigure}
\end{minipage}
\hspace{0.5cm}  % Adjust spacing between minipages
\caption{ Normal $N(s)$ and shear $H(s)$ tractions and bending moment $M(s)$ along the rod $s \in [0, L]$ for some tensions of the TCAMs $T_{11} \in [0, 20]$ and water free-stream velocity of $v_{\infty}= 0.2$ m/s.}
\label{Fig_NormalShearBendingStress}
\end{figure*}

The variation of the cross-sectional normal strain $\beta(s) \in \mathbb{R}$ along the arm is depicted in Figure~\ref{Fig_PlotSome_Beta} for various tensions $T_{11} \in [0, 20]$ $N$ under a water free-stream velocity of $v_{\infty}= 0.2$ $m/s$.
$\beta(s)$ represents the planar deformation of the cross-section in the direction of the standard unit director $\mathbf{b}(s) \in \mathbb{R}^2$. 
This deformation enables the arm's cross-section to adjust locally while preserving the local volume. This behavior aligns with the biomechanical principles observed in octopus arms, where tissue stiffness depends on incompressibility.
Fig.~\ref{Fig_PlotSome_Beta} indicates that $\beta(s)$ increases monotonically along the arm for all tensions $T_{11} \in [0, 20]$. 
%This means that $\beta(s)$ is greater in regions with smaller cross-sectional radii and continues to increase steadily at any fixed $s \in [0, L]$ as TCAM tensions $T_{11}$ increase.
To further explore arm mechanics, Fig.~\ref{Fig_NormalShearBendingStress} illustrates the variations in the normal $N(s)$ and shear $H(s)$ tractions, as well as the bending moment $M(s)$ along the arm for selected TCAM tensions. 
%it is insightful to examine variations in the normal $N(s)$ and shear $H(s)$ tractions, as well as the bending moment $M(s)$ defined in \eqref{Eq_BulkForceNHMConstitutive}, along the arm for selected TCAM tensions. 
%Figs.~\ref{Fig_PlotSome_NStress}, \ref{Fig_PlotSome_HStress}, and \ref{Fig_PlotSome_MStress} indicate an overall upward trend in these stresses and moment along the arm's length. 
%Similarly, within any cross-section at a fixed $s \in [0, L]$, these stresses and moment increase as TCAM tensions $T_{11} \in [0, 20]$ $N$ rise.
%
%%%%%%%%%%%% Figure %%%%%%%%%%%%%%%%%%%%%%
%
%%%%%%%%%%%%%%%%%%%%%%%%%%%%%%%%%%%%%%
\subsubsection{Validation}
\label{Sec_Validation}
\begin{figure}[t]
\centering
\begin{minipage}{0.48\textwidth}
\centering
\begin{subfigure}[t]{\textwidth}
\centering
\includegraphics[width=\textwidth, height=1.8in]{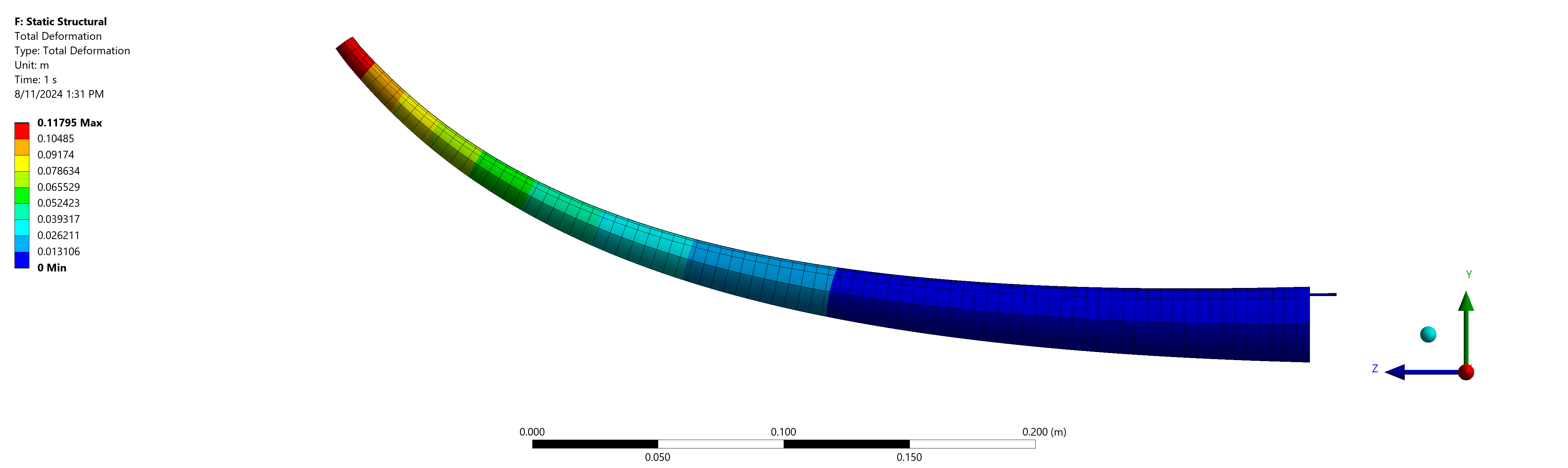}
\captionsetup{justification=centering}
\caption{Arm deformation for $v_{\infty}= 0.2$ $m/s$.}
\label{Plot_DeformedConfig_1WayFSI_V02}
\end{subfigure}
\end{minipage}
\hspace{0.04\textwidth}
\begin{minipage}{0.48\textwidth}
\centering
\begin{subfigure}[t]{\textwidth}
\centering
\includegraphics[width=\textwidth, height=1.8in]{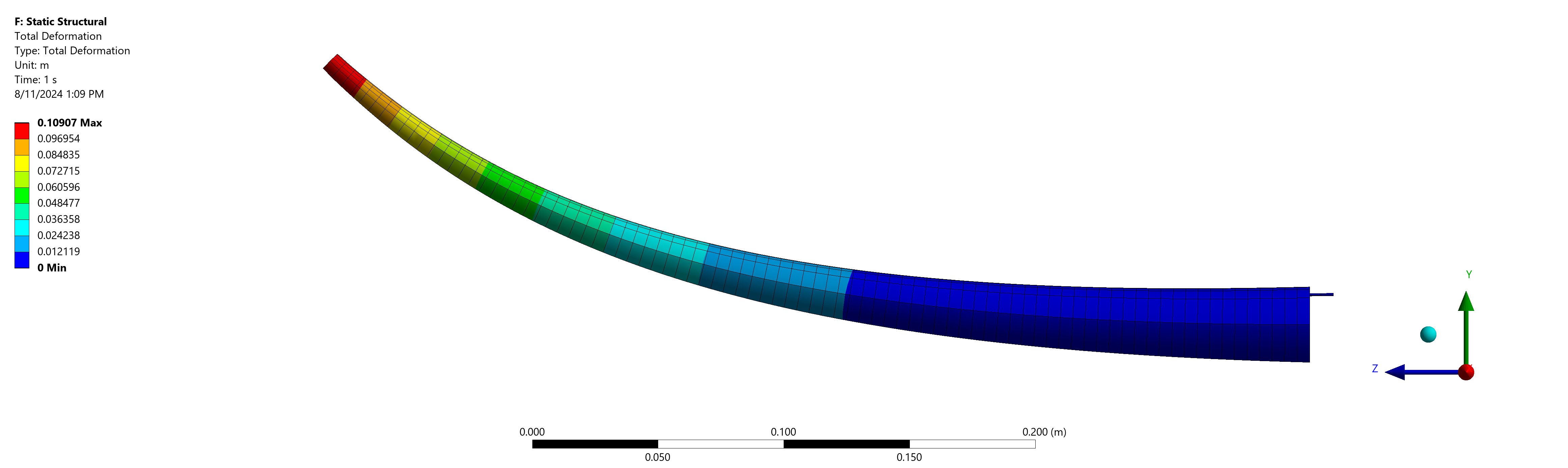}
\captionsetup{justification=centering}
\caption{Arm deformation for $v_{\infty}= 0.4$ $m/s$.}
\label{Plot_DeformedConfig_1WayFSI_V04}
\end{subfigure}
\end{minipage}
\caption{Deformations of the arm modeled in Ansys with FSI. }
\label{Plot_DeformedConfig_1WayFSI_V0204}
\end{figure}

\begin{figure}[t]
\centering
\begin{minipage}{0.48\textwidth}
\centering
\begin{subfigure}[t]{\textwidth}
\centering
\includegraphics[width=\textwidth, height=1.8in]{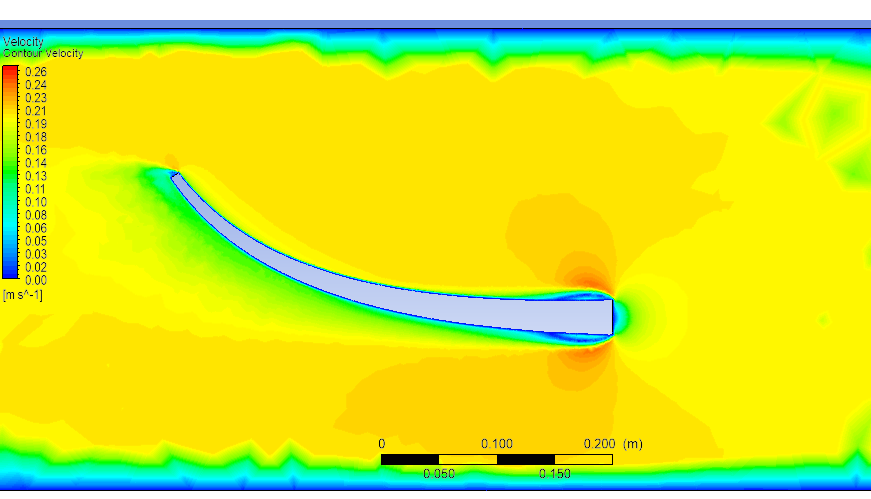}
\captionsetup{justification=centering}
\caption{Velocity field over the arm for $v_{\infty}= 0.2$ $m/s$.}
\label{Plot_Velocity_V02}
\end{subfigure}
\end{minipage}
\hspace{0.04\textwidth}
\begin{minipage}{0.48\textwidth}
\centering
\begin{subfigure}[t]{\textwidth}
\centering
\includegraphics[width=\textwidth, height=1.8in]{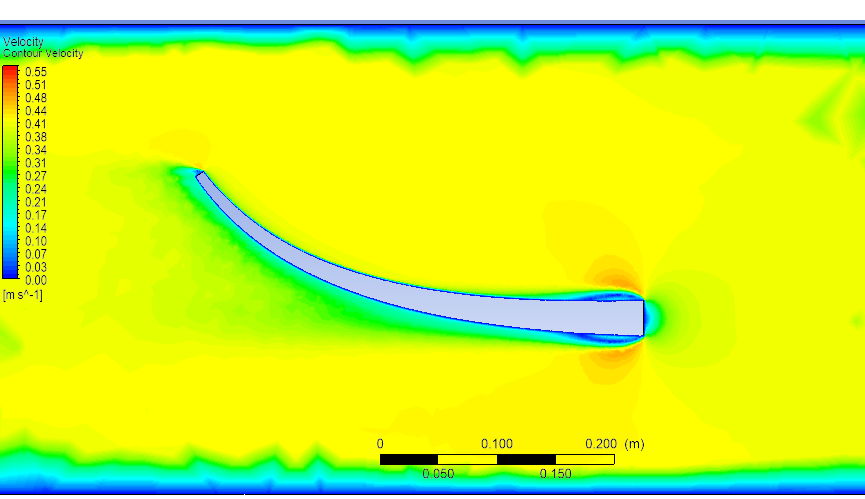}
\captionsetup{justification=centering}
\caption{Velocity field over the arm for $v_{\infty}= 0.4$ $m/s$.}
\label{Plot_Velocity_V04}
\end{subfigure}
\end{minipage}
\caption{Distribution of the velocity fields over the arm. }
\label{Plot_Velocity_V0204}
\end{figure}

\begin{figure}[t]
\centering
\begin{minipage}{0.48\textwidth}
\centering
\begin{subfigure}[t]{\textwidth}
\centering
\includegraphics[width=\textwidth, height=1.8in]{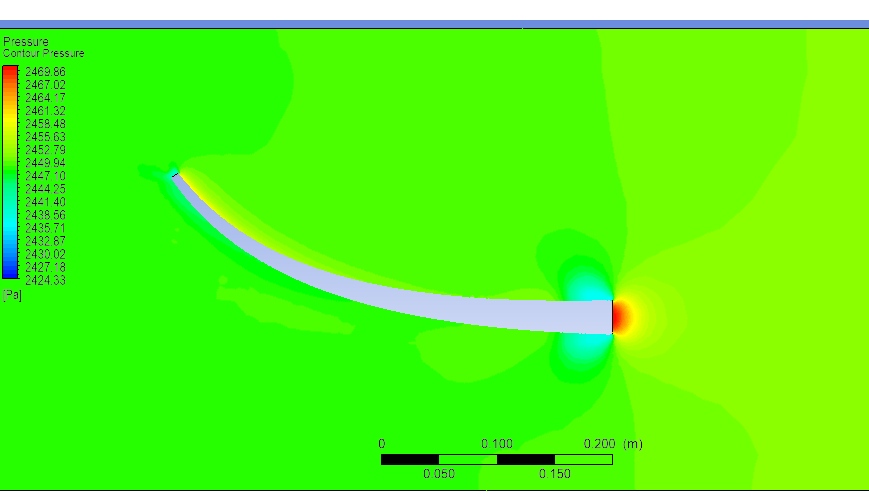}
\captionsetup{justification=centering}
\caption{Pressure field over the arm for $v_{\infty}= 0.2$ $m/s$.}
\label{Plot_Pressure_V02}
\end{subfigure}
\end{minipage}
\hspace{0.04\textwidth}
\begin{minipage}{0.48\textwidth}
\centering
\begin{subfigure}[t]{\textwidth}
\centering
\includegraphics[width=\textwidth, height=1.8in]{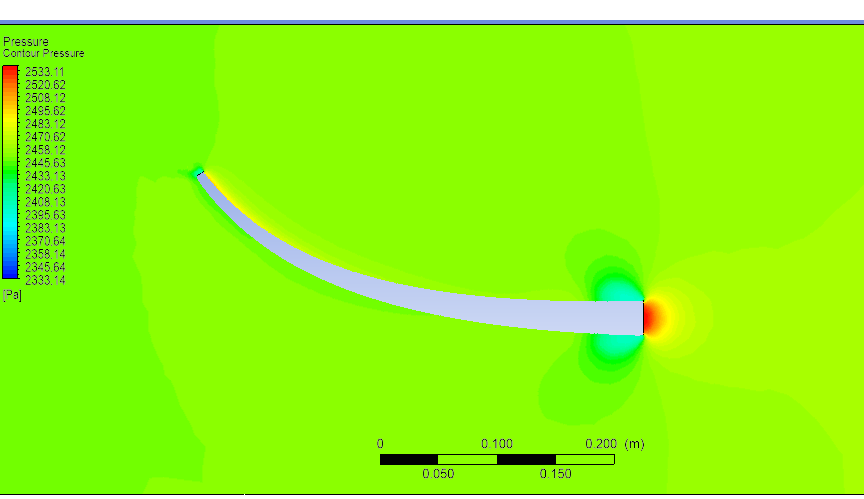}
\captionsetup{justification=centering}
\caption{Pressure field over the arm for $v_{\infty}= 0.4$ $m/s$.}
\label{Plot_Pressure_V04}
\end{subfigure}
\end{minipage}
\caption{Distribution of the pressure fields over the arm. }
\label{Plot_Pressure_V0204}
\end{figure}

To validate the Cosserat-based model for the octopus-like arm in water, Ansys simulations were conducted under similar conditions.
Due to the limitations of beam elements in Ansys for fluid-structure interaction (FSI) analysis, three-dimensional (3D) solid elements were utilized, and an FSI approach was employed to accurately model the interaction between the soft arm and the steady-state water flow.
Fig.~\ref{Plot_DeformedConfig_1WayFSI_V0204} shows the deformations of the arm modeled in Ansys with FSI, while Figs.~\ref{Plot_Velocity_V0204} and \ref{Plot_Pressure_V0204} respectively present the distribution of velocity and pressure fields over the arm. 
%
%%%%%%%%%%%%% Figure %%%%%%%%%%%%%%%%%%%%%
%
%%%%%%%%%%%%%%%%%%%%%%%%%%%%%%%%%%%%%%
%%%%%%%%%%%%% Figure %%%%%%%%%%%%%%%%%%%%%
%
%%%%%%%%%%%%%%%%%%%%%%%%%%%%%%%%%%%%%%
%%%%%%%%%%%% Figure %%%%%%%%%%%%%%%%%%%%%%
%
%%%%%%%%%%%%%%%%%%%%%%%%%%%%%%%%%%%%%%
The Ansys simulation results generally show strong agreement with the Cosserat-based model, particularly in terms of the maximum deformations at the arm tip, confirming the model's accuracy. 
However, some discrepancies are observed at the arm tip, where the Ansys results differ in curvature compared to the Cosserat-based results. 
This divergence arises because the extended Cosserat theory models the arm as a rod, effectively capturing the complex behavior of the arm’s curvature and the orientation of its cross-section, which are particularly critical at the tip where curvature is more pronounced. 
In contrast, the 3D solid elements in Ansys do not explicitly model the arm's curvature as a rod or the orientation of the cross-section, leading to variations in predicted curvature and overall deformation, especially in regions with higher curvature. This difference in modeling approaches explains the observed discrepancies, particularly at the arm’s tip.
%
%%%%%%%%%%%%%%%% Conclusions %%%%%%%%%%%%%%%%%%%
\section{Conclusions}
\label{Sec_Conclusions}
This study explored the use of recently developed TCAMs to actuate and replicate the bending motion of an octopus-like soft robot arm underwater. 
A quasi-static continuum model was developed to examine the effects of hydrostatic and dynamic forces from steady-state fluid flow on the arm's bending motion. 
The extended Cosserat theory of rods was used to model the soft robot arm as a continuum robot, allowing for rigid rotation of the cross-section without constraining it to be locally normal to the rod's direction. 
Unlike the standard Cosserat theory, this approach included planar deformation of the cross-section, introducing a new normal strain to describe in-plane deformation. 
A constitutive model for the octopus arm material was also proposed to capture its characteristic behavior.
%
%%%%%%%%%%%%%%%%%%%%%%%%%%%
%\newpage
\bibliographystyle{plain} 
\bibliography{Main3}

%\clearpage

\end{document}